\documentclass{article}

\usepackage{arxiv}

\usepackage[utf8]{inputenc} % allow utf-8 input
\usepackage[T1]{fontenc}    % use 8-bit T1 fonts
\usepackage{hyperref}       % hyperlinks
\usepackage{url}            % simple URL typesetting
\usepackage{booktabs}       % professional-quality tables
\usepackage{amsfonts}       % blackboard math symbols
\usepackage{nicefrac}       % compact symbols for 1/2, etc.
\usepackage{microtype}      % microtypography
\usepackage{lipsum}
\usepackage{graphicx}

\newcommand{\ra}[1]{\renewcommand{\arraystretch}{#1}}
	\usepackage[ruled]{algorithm2e}
	\SetKwInput{KwIn}{Input}
	\SetKwInput{KwOut}{Output}
	\SetKwInput{KwPar}{Parameters}
	\SetKwInput{KwIns}{Instructions}

\usepackage{amsmath}
	\usepackage{amsthm}
	\usepackage{float}
	\usepackage{graphicx}
	\usepackage{amssymb}
	\usepackage{diagbox}
	\usepackage{booktabs}
	\usepackage{mathrsfs}
	\usepackage{color}
	\usepackage{bm}
	\usepackage{caption}
	\usepackage{subfig}
	\usepackage{enumerate}
	\usepackage{multicol}
	\usepackage{multirow}
	\usepackage{dsfont}
	\usepackage{booktabs}
	\usepackage{sidecap}
	\usepackage{cite}

	\newtheorem{remark}{Remark}
	
\graphicspath{ {./images/} }

\title{Extended Object Tracking and Classification based on Linear Splines}

\author{
 Matteo Tesori \\
  Università degli Studi di Firenze\\
  \texttt{matteo.tesori@unifi.it} \\
   \And
 Giorgio Battistelli \\
  Università degli Studi di Firenze\\
  \texttt{giorgio.battistelli@unifi.it} \\
  \And
 Luigi Chisci \\
  Università degli Studi di Firenze\\
  \texttt{luigi.chisci@unifi.it} 
}

\begin{document}
\maketitle
\begin{abstract}
This paper introduces a framework based on linear splines for 2-dimensional extended object tracking and classification. Unlike state of the art models, linear splines allow to represent extended objects whose contour is an arbitrarily complex curve. 
 An exact likelihood is derived for the case
 in which noisy measurements can be scattered from any point on the contour of the extended object, while an approximate Monte Carlo likelihood is provided for the case wherein scattering points can be anywhere, i.e. inside or on the contour, on the object surface.
 Exploiting such likelihood to measure how well the observed data fit a given shape, a suitable estimator is developed. The proposed estimator models the extended object in terms of a kinematic state, providing
	object position and orientation, along with a shape vector, characterizing object contour and surface.
	The kinematic state is estimated via a nonlinear Kalman filter, while the shape vector is estimated via a Bayesian classifier so that classification is implicitly solved during shape estimation.
	Numerical experiments are provided to assess, compared to state of the art extended object estimators, the effectiveness of the proposed one.
\end{abstract}

% keywords can be removed

\section{Introduction}
As cutting-edge high-resolution sensors become more prevalent, the landscape of object tracking is undergoing a notable transformation, placing a growing emphasis on so called \textit{extended objects} \cite{granstrom2017extended, granstrom2014gmPHD,  granstrom2014newprediction, granstrom2012phdx, granstrom2013spawning, beard2016multipleext, beard2015glmbx, vivone2016multistatic}. These differ from  conventional \textit{point objects} in that they span across multiple resolution cells of the sensor. Due to this fact, extended objects tend to generate multiple measurements during each sensor scan, thus enabling the estimation of their shape beside their kinematic state.

\begin{figure}
\centering
\includegraphics[scale=1.15]{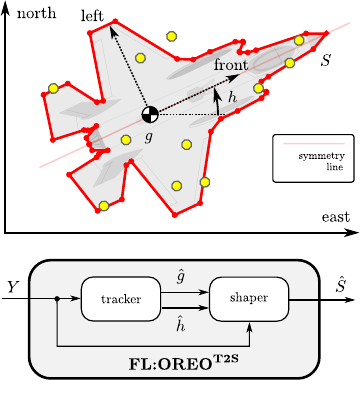}
\caption{Illustration of the extended object model and architecture of the proposed estimator, referred to as \textit{Fitting Lambda:Omicron Recursive estimator for Extended Objects} (FL:OREO). The object is represented by a sequence of vertices (red dots) collected in a vector $S$. The shape vector $S$ is recursively estimated from a dataset $Y$ composed by a random set of measurements scattered around the object (yellow dots).
The vector $S$ is estimated in two steps: (1) a \textit{tracker} estimates the position $g$ and the heading $h$; (2) a \textit{shaper} generates the estimate $\hat{S}$.
	This estimation scheme is referred to as \textit{Track To Shape} (T2S).
}
\label{fig:sexypic}
\end{figure}

The additional information provided by the object shape can be exploited for classification purposes, i.e. to recognize which class the tracked object belongs to. This fact suggests that extended object tracking and object classification are closely related to each other and, not surprisingly,  classifiers based on extended object trackers have been developed \cite{cao2018esm, cheng2024LMB, wang2019RMMBF, wang2024starconv}. 
	
This work proposes a solution for extended object tracking and classification based on the idea that in certain circumstances shape estimation can be regarded as a classification problem. In this respect, a central contribution of this work is the development of a shape classifier for arbitrarily complex object shapes, wherein the main challenge consists in the definition of an effective likelihood model.
To achieve this goal, the contours of planar objects are modeled as piece-wise linear curves, referred through the paper as \textit{linear splines}, or polygonal chains, or polygonal curves or polygonal lines. 
	
It is worth to point out that the proposed approach does not generate the estimated shape ex-novo like conventional extended object estimators do, but rather just selects the best shape (in the maximum a-posteriori sense) within a finite set of possible  known shapes. Clearly, this poses limits in the estimation of general shapes.
On the other hand, the proposed solution can effectively be employed in scenarios wherein there is good prior knowledge about the types of objects that have to be tracked (e.g. in air and vehicular surveillance scenarios). Moreover, the proposed solution can be employed in tandem with conventional extended object estimators: first, with moderate computational effort, the classification-based estimator of this paper can be used to generate an initial ``raw'' estimate; then, if the ``raw'' estimate is considered not acceptable, a conventional extended object estimator can be employed to refine it.
	
The rest of the paper is organized as follows: the remaining part of section I briefly reviews the current state of the art in extended object tracking and classification; section II introduces the linear spline framework; section III presents parametric simulations comparing the proposed estimators with state of the art estimators; section IV summarizes the main results and provides some direction for future developments.

\subsection{State of the art}
	
Shape models for extended object tracking fall into two main categories: \textit{parametric} and \textit{non-parametric}.
	
Parametric models operate under the assumption that the object contour is specified by a pre-fixed number of parameters. For instance, common models in this class assume that the object contour is either the perimeter of a rectangle \cite{granstrom2011rect, granstrom2014phdrect, petrovskaya2008model} or an ellipse, i.e. two curves fully characterized by a position vector, an orientation angle and two semi-axis lengths (totaling five parameters).  A more general model that encompasses both is the superellipse \cite{baur2024, Gao2024, Yurdakul2024}, which describes the perimeter of the contour curve using a third parameter to control convexity. This parameter allows the shape to vary from rectangular to elliptic and even to cross-shaped contours. The leading model in this category is the \textit{random matrix} one, which makes use of ellipses to represent object shapes \cite{koch2008bayesian, feldmann2011tracking, orguner2012variational, lan2014maneuverRMM, yang2019mem}.

Non-parametric models do not assume that the object contour can be described by a pre-fixed number of parameters. These models offer greater flexibility compared to parametric ones, but this advantage comes with solutions typically involving larger computational cost. A class of non-parametric models is the one of \textit{Random Hypersurfaces} (RH), where the main idea is to represent the object contour with a continuous radial function \cite{baum2014RHM, baum2009RHM, baum2011shape}. In their original forms, RH estimators perform shape estimation by estimating an arbitrary number of coefficients of the Fourier expansion relative to the radial function.
An alternative strategy to face shape estimation is based on \textit{Gaussian Processes} (GP), where the radial function is discretized with arbitrary resolution and shape estimation consists of estimating the values of the radial function at the discretization points \cite{wahlstrom2015GP, kumru20183DGPM}. RH models assume that the object shape is star-convex, an assumption that does not hold in some particular applications like, e.g.,  air surveillance. To mitigate this constraint, a possible approach is to model the object shape as a deformation of an ellipse. In \cite{cao2021eda}, shape estimation is carried out by estimating the positions of an arbitrary number of points lying on the boundary of the deformed ellipse.
	
Concerning extended object classification, a common strategy is to classify the object on the basis of the output of an extended object tracker \cite{tuncer2018, Cao2022ExtendedOT, lan2014NRMM}. A different standpoint is to actively employ the classifier to generate the object shape. In \cite{hoher2023eotshape}, three distinct object classes based on primitive shapes (ellipses, triangles, and rectangles) are considered. Then, the shape estimate is generated by selecting the most probable primitive shape. A different model is presented in \cite{sun2014support}, wherein support functions are utilized to represent primitive shapes which are subsequently combined together to generate the shape estimate.

\section{Basic extended object models}
	\subsection{General contour model}
	
	To model the set of resolution cells occupied by an extended object, consider a non-self-intersecting closed curve $\psi:\alpha\in[0,1]\mapsto \mathbb{R}^2$. More precisely, the 
 curve $\psi(\alpha)$ outlines the object contour through the set of points
	\begin{equation}
	c\triangleq \{\psi(\alpha)\}_{\alpha\in[0,1]},
	\end{equation}
	while the resolution cells occupied by the object are identified by the surface area encircled by the contour $c$ identified by $\psi(\alpha)$. The set of points that are inside the region surrounded by $\psi(\alpha)$ is shortly denoted as $\mathring{c}$ or $\mathrm{int}\,c$.

	Note that at this initial stage, the curve $\psi(\alpha)$ provides, somewhat, a joint description of the object ``location'', ``orientation'' and ``shape''.
	
	A possible way to define the object ``location'' is provided by the \textit{barycenter} $g\in\mathbb{R}^2$. If $\psi(\alpha)$ is smooth almost everywhere, then the barycenter $g$ is defined as  
	\begin{equation}
	g \triangleq \frac{\int_0^1 \psi(\alpha) \lVert \dot{\psi}(\alpha)\rVert \,\mathrm{d}\alpha}{\int_0^1 \lVert \dot{\psi}(\beta) \rVert \,\mathrm{d}\beta}
	\end{equation}
	where $\dot{\psi}(\alpha)\triangleq \frac{\mathrm{d}\psi}{\mathrm{d}\alpha}(\alpha)$ is the tangent vector at $\alpha$ and $\lVert \zeta\rVert$ denotes the Euclidean norm of the generic vector $\zeta$. Note that both $\psi(\alpha)$ and $g$ are expressed with respect to the world reference system or, more shortly, in \textit{world coordinates}. 
	
	Assume that the contour curve $c(\alpha)$ is symmetric with respect to a line passing through the barycenter $g$. Let $h\in[0,2\pi)$ be the counter-clockwise angle between the symmetry line and the east semi-axis of the world reference system. Then the \textit{heading} angle $h$ provides a possible definition for the object ``orientation''. Note that most vehicles are characterized by a symmetry line, which often coincides with the vehicle longitudinal axis.
	
	Based on $g$ and $h$, the local reference system can be defined as the one centered in $g$ and rotated around $g$ by the angle $h$. Such reference system is usually called \textit{barycentric} and the curve
	\begin{align}
	\label{eq:barCurve}
	\overline{\psi}(\alpha)&= U(h)' \, (\psi(\alpha)-g) 
	\qquad  \alpha \in[0,1] \\
	U(h)&\triangleq \left[\begin{array}{cc}
	\cos h & -\sin h \\
	\sin h & \phantom{-}\cos h
	\end{array}\right]
	\end{align}
	is nothing but the expression in barycentric coordinates of the original contour curve $\psi(\alpha)$.
	
	While the object location and orientation are specified by $g$ and $h$, the curve $\overline{\psi}(\alpha)$ uniquely characterizes
 the object shape, since the object contour or surface are encoded by the sets of points
	\begin{equation}
	\overline{c}\triangleq \left\{\overline{\psi}(\alpha)\right\}_{\alpha\in[0,1]}\qquad\mathring{\overline{c}}\triangleq \mathrm{int} \,\overline{c}.
	\end{equation}
	According to (\ref{eq:barCurve}), the object shape $\overline{\psi}(\alpha)$ can be combined with the location $g$ and heading $h$ as
	\begin{equation}
	\label{eq:barc2c}
	\psi(\alpha)\triangleq U(h)\,\overline{\psi}(\alpha)+g 
	\qquad  \alpha \in[0,1]
	\end{equation} 
	to obtain the world coordinate representation $\psi(\alpha)$ of the object shape $\overline{\psi}(\alpha)$ which, as already mentioned, provides a joint description of object shape, position and orientation.
	
	\begin{remark}If the object contour $c$ outlines a star-convex surface, then $\psi(\alpha)$ reduces to a radial function. In this sense, the curve $\psi(\alpha)$ can be regarded as a generalized radial function able to represent shapes that are not necessarily star-convex. Naturally, this is also true for $\overline{c}$ and $\overline{\psi}(\alpha)$.
	\end{remark}

\subsection{Measurement models}
	An extended object can be seen as a scattering domain, where each point of such domain can be detected or not by the sensor. When a scattering point is detected, its observation is corrupted by noise.
	Accordingly, the following model is considered
\begin{align}
	\label{eq:measModel}
	y&=z+v\\
	z&\sim p_z(z;\psi)\\
	v&\sim p_v(v)
	\end{align}
	where $y$ is a measurement observed by the sensor, $z$ is the scattering point location and $v$ is the measurement noise. Moreover, $z$ and $v$ are assumed to be independent random vectors distributed according to the probability densities $p_z(z;\psi)$ and, respectively, $p_v(v)$. Note that the scattering density $p_z(z;\psi)$ also depends on the contour curve $\psi(\alpha)$, since such curve characterizes the scattering domain associated to the extended object.

	Due to (\ref{eq:measModel}) and by independence, the single measurement likelihood is given by the following convolution.
	\begin{equation}
	\label{eq:conv}
	\mathcal{L}\left(y|\psi\right)=\int p_v(y-z)\,p_z(z;\psi)\,\,\mathrm{d}z.
	\end{equation}

	A common choice for the measurement noise distribution is the Gaussian one
	\begin{equation}
	\label{eq:pv}
	p_v(v)\triangleq \mathcal{N}\left(v;0,R\right)
	\end{equation}
	where $R=R'>0$ and 
	\begin{align}
	\mathcal{N}(x; \mu, \Sigma)\triangleq \frac{\exp
 \left( -\frac{1}{2}
 (x-\mu)' \Sigma^{-1} (x-\mu) \right)}
 {\sqrt{\mathrm{det}\, (2 \pi \Sigma)}}
	\end{align}
	denotes the Gaussian density with mean $\mu$ and covariance $\Sigma$. 
	
For what concerns the 
%definition of the 
scatter density $p_z(z;\psi)$, the following two models, 
named \textit{contour} and \textit{surface} model \cite{wahlstrom2015GP}, are considered.
	\begin{enumerate}
	\item \textit{Contour model}: the scatter point $z$ is assumed to be uniformly distributed over the object contour $c$. In this case the scattering density of $z$ is denoted as $\mathcal{U}(z; c)$, while the resulting measurement likelihood is denoted as $\mathcal{L}(y|c)$. As it will be shown, if the object contour $\psi(\alpha)$ is a linear spline then $\mathcal{L}(y|c)$ admits an exact closed form.
	
	\item \textit{Surface model}: the scatter point $z$ is assumed to be uniformly distributed over the object surface $\mathring{c}$.
	In this case the scattering density of $z$ is denoted as $\mathcal{U}(z; \mathring{c})$, while the resulting measurement likelihood is denoted as $\mathcal{L}(y|\mathring{c})$. Unfortunately, an exact closed-form expression for $\mathcal{L}(y|\mathring{c})$ is not available. However, as it will be shown, if the object contour $\psi(\alpha)$ is a linear spline then a Monte Carlo approximation of $\mathcal{L}(y|\mathring{c})$ can be obtained.
	\end{enumerate}
	
		In what follows, $\mathcal{L}(y|\psi)$ denotes the generic measurement likelihood that can be either $\mathcal{L}(y|c)$ or $\mathcal{L}(y|\mathring{c})$.
		
	\begin{remark}
	Consider the following circular curve
	\begin{equation*}
	\psi(\alpha)\triangleq \rho\,\left[\begin{array}{cc}
	\cos \alpha & \sin \alpha
	\end{array}\right]'+z_g \qquad \alpha \in[0,1]
	\end{equation*}
	where $\rho>0$ is a given radius and $z_g\in\mathbb{R}^2$ a given position in the plane. If $\rho\to 0$, the contour $c$ and the surface $\mathring{c}$ of the curve $\psi(\alpha)$ collapse over the single point $z_g$. In this particular case the scattering densities $\mathcal{U}(z;c)$ and $\mathcal{U}(z; \mathring{c})$ reduce to a Dirac delta centered in $z_g$, meaning that $z=z_g$ almost surely. Consequently, 
	\begin{equation}
	\label{eq:pointlik}
	\lim_{\rho \to 0} \mathcal{L}\left(y|\psi\right) = \mathcal{N}(y; z_g, R)
	\end{equation}
	which is the conventional likelihood used in point object tracking.
	Equation (\ref{eq:pointlik}) shows that the proposed measurement model (\ref{eq:conv}) generalizes the usual one employed in point object tracking, i.e.
	\begin{align}
	y&=z_g+v\\
	v&\sim\mathcal{N}(v;0,R)
	\end{align}
	where $z_g$ is the true position of the tracked object.
	\end{remark}

	\begin{remark} Let $\rho_{\max}\triangleq \max_{z\in \overline{c}} \lVert z\rVert$ be the radius of the minimum area circle circumscribed to the scattering domains $c$ and $\mathring{c}$, referred to as the object \textup{outer radius}. Consider the special case where the measurement noise is isotropic, i.e. $R\triangleq \sigma^2 I$ for a given standard deviation $\sigma > 0$.  
	Let $\sigma \gg \rho_{\max}$.  Then, for all $z\in c$ or for all $z\in \mathring{c}$, 
 the event $\lVert v\rVert\gg z$ is almost sure,  so that $y\approx v$. Consequently,
	\begin{equation}
	\label{eq:isolik2}
	\lim_{\sigma \to \infty}\mathcal{L}(y|\psi)=\mathcal{N}(y;0,R)
	\end{equation}
	which does not depend on the object contour curve $\psi(\alpha)$. Hence, as $\sigma\to \infty$, different scattering domains become indistinguishable. 
	In short, an extended object is \textup{unresolvable} whenever its likelihood  is given by (\ref{eq:isolik2}). Due to (\ref{eq:isolik2}), $\sigma$ controls the ability of the likelihood model (\ref{eq:conv}) to distinguish different objects.
	\end{remark}

\subsection{Contour model}
	Since the tracked object is extended, at each scan the sensor can detect a number $m\in\mathbb{N}$ of measurements. Such measurements can be collected in the dataset
	\begin{equation}
	\label{data}
	Y\triangleq\left\{y^{(j)}\right\}_{j=1}^{m}.
	\end{equation}
	Let $m>0$. By assuming that each measurement $y^{(j)}$ is independent of the other ones, it follows that the dataset likelihood factorizes into the product of the $m$ measurement likelihoods (\ref{eq:conv}) relative to each individual measurement $y^{(j)}$. 
 Conversely, if $m=0$, the dataset likelihood reduces to the probability that the dataset is empty. Consequently, a first model for the dataset likelihood is
	\begin{equation}
	\label{eq:exactlik}
	\begin{aligned}
	&\mathcal{L}\left(Y|c, m\right)\triangleq\begin{cases}
	\hfil\prod_{j=1}^m \mathcal{L}\left(y^{(j)}|c\right) & \text{if } m>0\\
	\hfil\mathbb{P}(Y=\varnothing) & \text{otherwise .}
	\end{cases}
	\end{aligned}
	\end{equation}
	
\begin{figure*}[h]
	\centering
	\includegraphics[scale=0.33]{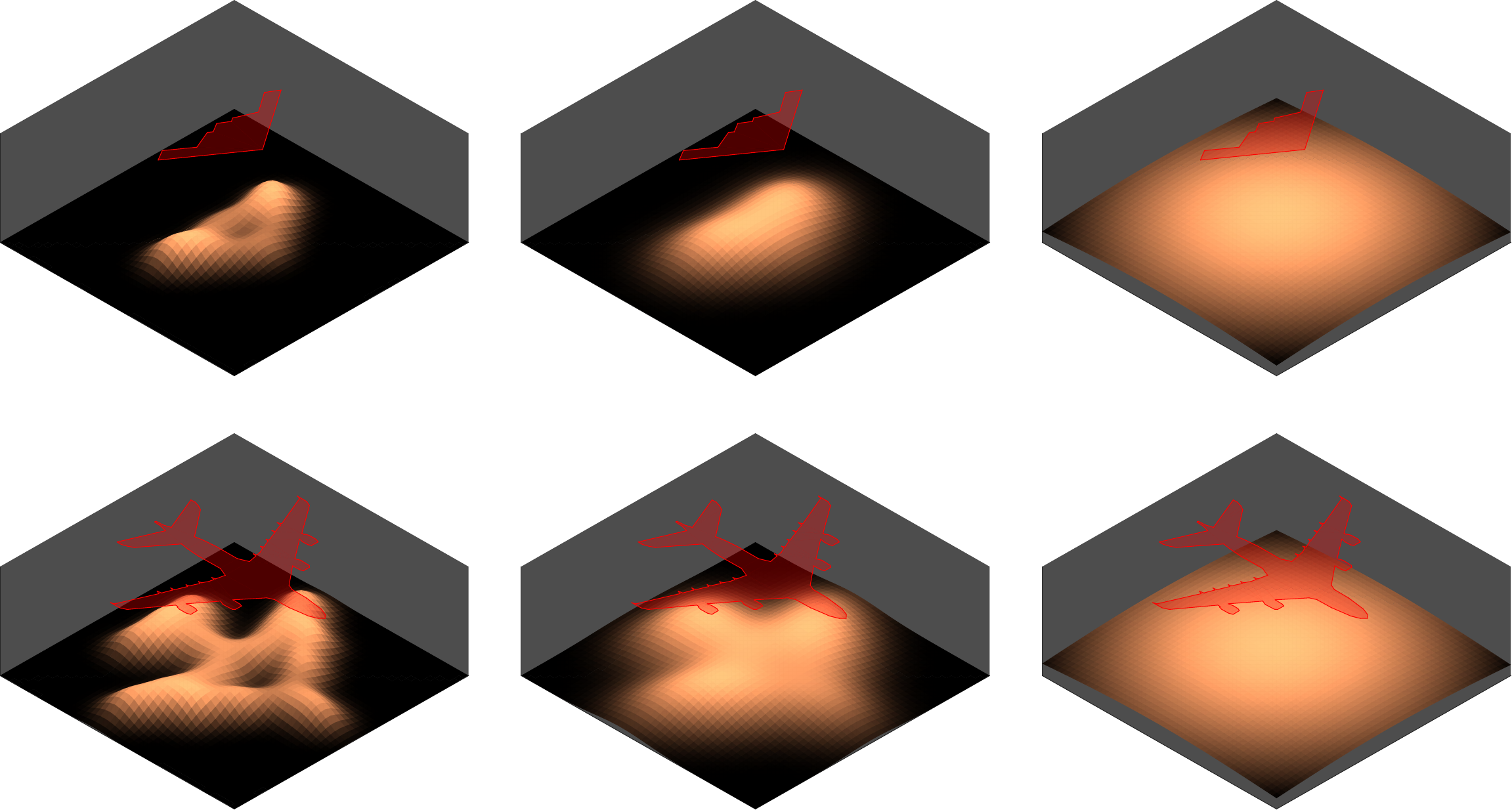}
	\caption{Visualization of the single measurement likelihood (\ref{eq:exactlik})
	$\mathcal{L}:y\in[-\mathrm{FoV},\mathrm{FoV}]^2\mapsto \mathbb{R}_{\geq 0},\,\,\mathrm{FoV}\triangleq 50\,\,[\mathrm{m}]$, 
	for two different scattering domains (top row: contour of a stealth bomber with outer radius $\rho_{\max}\approx 27\,[\mathrm{m}]$; bottom row: contour of an airliner with outer radius $\rho_{\max}\approx 42\,[\mathrm{m}]$) and three different values for the noise 
	covariance $R\triangleq \sigma^2I$ (left column: $\sigma\triangleq 5\,\,[\mathrm{m}]$; center column: $\sigma\triangleq 10\,\,[\mathrm{m}]$; right column: $\sigma\triangleq 50\,\,[\mathrm{m}]$).  
	For $\sigma\triangleq 50\,[\mathrm{m}]$, both domains are unresolvable, i.e. cannot be distinguished in terms of the dataset likelihood (\ref{eq:exactlik}).
	}
	\end{figure*}
	
	In many tracking applications, e.g. radar tracking, the dataset cardinality $m$ is a time-varying quantity whose dynamics cannot be expressed by means of simple and deterministic models. 
	A possible approach to capture the fluctuations in the value of the dataset cardinality is to regard it as a random variable following some discrete distribution $p_c(m):\{0,1,,\dots\}\mapsto [0,1]$ that, in general, depends on the object contour $c$. Combining (\ref{eq:exactlik}) with $p_c(m)$ leads to the refined model
	\begin{equation}
	\label{eq:exactDatasetlik}
	\begin{aligned}
	\mathcal{L}\left(Y|c\right)\triangleq\begin{cases}
	\hfil p_c(m)\,\prod_{j=1}^m \mathcal{L}\left(y^{(j)}|c\right) & \text{if } m>0\\
	\hfil p_c(0) & \text{otherwise}
	\end{cases}
	\end{aligned}
	\end{equation}	
	where 
 the void probability $\mathbb{P}(Y=\varnothing)$ in \eqref{eq:exactlik} actually coincides with $p_c(0)$ in \eqref{eq:exactDatasetlik}. It is worth to point out that (\ref{eq:exactDatasetlik}) simultaneously takes into account: the dispersion of $Y$ with respect to the object contour $c$; 
 the cardinality of $Y$ with respect to the object contour $c$. 
	
	Thanks to the cardinality factor $p_c(m)$, (\ref{eq:exactDatasetlik}) can distinguish different objects even when their scattering domains are unfocused. To see why, let $R\triangleq \sigma^2 I$ and $m>0$; then (\ref{eq:exactDatasetlik}), under intense measurement noise, reduces to
	\begin{equation}
	\label{eq:isoexactDatasetlik}
	\lim_{\sigma \to \infty} \mathcal{L}(Y|c)=p_c(m) \prod_{j=1}^{m} \mathcal{N}\left(y^{(j)};0, R\right)
	\end{equation}
	which, unlike (\ref{eq:exactlik}), in general retains the dependence on the object contour $c$. Equation (\ref{eq:isoexactDatasetlik}) shows that (\ref{eq:exactDatasetlik}) is more resilient to measurement noise than (\ref{eq:exactlik}).
	
A possible model for the cardinality distribution $p_c(m)$ is given by the binomial distribution	
	\begin{equation}
	\label{eq:binom}
	\mathcal{B}\left(m; c\right)\triangleq
	\binom{\overline{\mu}_{c}}{m}\,\overline{\pi}_{c}^m\,\left(1-\overline{\pi}_{c}\right)^{\overline{\mu}_{c}-m}.
	\end{equation}
	where the binomial parameters $\overline{\mu}_{c}\in\mathbb{N}$, $\overline{\pi}_{c}\in[0,1]$ depend on the object contour $c$, meaning that the specific value $m$ of the dataset cardinality can provide useful information to distinguish different objects.
	
	A possible model for the number of trials $\overline{\mu}_{c}$ is 
	 \begin{equation}\overline{\mu}_{c}\triangleq \left\lfloor \frac{|c|}{\overline{r}}\right\rceil
	\end{equation}
	where: $\lfloor x\rceil$ is the rounded value of the generic real $x$;  $|c|$ is the contour length (expressed in $[\mathrm{m}]$), i.e.
	\begin{equation}
	|c| \triangleq \int_0^1 \lVert \dot{\psi}(\alpha)\rVert\, \mathrm{d}\alpha;
	\end{equation}
	$\overline{r}>0$ is a parameter (expressed in $[\mathrm{m}]$) somewhat connected to the sensor resolution. 
	
	A possible model for the success probability $\overline{\pi}_{c}$ is 
	\begin{equation}
	\label{eq:pdc}
	\overline{\pi}_{c} \triangleq \overline{\Gamma}_c\,\overline{\eta}\\
	\end{equation}
	where $\overline{\eta}\in[0,1]$ and, for ${\Gamma}_c:\mathbb{R}\mapsto [0,1]$,
	\begin{equation}
	\overline{\Gamma}_c \triangleq \frac{1}{|c|}\int_0^1 \Gamma_c(\alpha)\,\lVert \dot{\psi}(\alpha)\rVert \, \mathrm{d}\alpha.
	\end{equation}
	Model (\ref{eq:pdc}) takes inspiration from radar applications: the second factor $\overline{\eta}$ is a parameter that models the capacity of the sensor to illuminate the tracked object with an electromagnetic wave; the first factor $\overline{\Gamma}_c$ models the predisposition of the tracked object to reflect back to the sensor such electromagnetic wave. Similarly, the surrogate weighting function $\Gamma_c$ describes the predisposition of the generic point $z\in c$ to reflect back the electromagnetic wave generated by the sensor. Due to this interpretation, $\overline{\Gamma}_c$, $\Gamma_c$ and $\overline{\eta}$ are referred to as \textit{(global) reflectivity}, \textit{(local) reflectivity} and \textit{(sensor) lighting power}.

\subsection{Surface model}
Similarly to the contour model, the dataset model considered in the surface case is
 \begin{equation}
	\label{eq:exactDatasetlik2}
	\begin{aligned}
	\mathcal{L}\left(Y|\mathring{c}\right)\triangleq\begin{cases}
	\hfil p_{\mathring{c}}(m)\,\prod_{j=1}^m \mathcal{L}\left(y^{(j)}|\mathring{c}\right) & \text{if } m>0\\
	\hfil p_{\mathring{c}}(0) & \text{otherwise}
	\end{cases}
	\end{aligned}
	\end{equation}	
where $p_{\mathring{c}}(m):\{0,1,2,\dots\}\mapsto [0,1]$ is some probability distribution depending on the object surface $\mathring{c}$ and modeling the fluctuations of the dataset cardinality. 

Following the same rational of the contour case, a possible model for $p_{\mathring{c}}(m)$ is the binomial one given by
\begin{align}
	\label{eq:binom}
	\mathcal{B}\left(m; \mathring{c}\right)\triangleq
	\binom{\overline{\mu}_{\mathring{c}}}{m}\,\overline{\pi}_{\mathring{c}}^m\,\left(1-\overline{\pi}_{\mathring{c}}\right)^{\overline{\mu}_{\mathring{c}}-m}
	\end{align}
	where, by introducing the contour area (expressed in $[\mathrm{m}^2]$) as
\begin{equation}
|\mathring{c}|\triangleq \int_{\mathring{c}} \mathrm{d}z
,\end{equation}	
the binomial parameters $\overline{\mu}_{\mathring{c}}\in\mathbb{N}$, $\overline{\pi}_{\mathring{c}}\in[0,1]$ are
	\begin{align}
	\overline{\mu}_{\mathring{c}} &\triangleq \left\lfloor \frac{|\mathring{c}|}{\mathring{\overline{r}}}\right\rceil \\
	\overline{\pi}_{\mathring{c}} &\triangleq \overline{\Gamma}_{\mathring{c}} \,\overline{\eta}\\
	\overline{\Gamma}_{\mathring{c}} &\triangleq \frac{1}{|\mathring{c}|}\int_{\mathring{c}} \Gamma_{\mathring{c}}(z)\, \mathrm{d}z.
	\end{align}
	
The parameter $\mathring{\overline{r}}>0$ (expressed in $[\mathrm{m}^2]$) represents the sensor resolution, while $\overline{\mu}_{\mathring{c}}$ represents the number of resolution cells occupied by the object. The parameter $\overline{\Gamma}_{\mathring{c}}$, which, likewise the contour case, has the meaning of average reflectivity, is obtained by some local reflectivty $\Gamma_{\mathring{c}}:\mathbb{R}^2 \mapsto [0,1]$. 
	
		\begin{remark}
	
	Except for a scaling factor equal to $|Y|!$, the refined models (\ref{eq:exactDatasetlik}), (\ref{eq:exactDatasetlik2})  represent the multi-object density of an \textup{Independent and Identically Distributed Cluster \cite{mahler2014advances}}. From this point of view, the dataset $Y$ can be seen as a \textup{Random Finite Set (RFS)} with arbitrary cardinality distribution $p_c(m)$ or $p_{\mathring{c}}(m)$ and arbitrary spatial density $\mathcal{L}\left(y|c\right)$ or $\mathcal{L}\left(y|\mathring{c}\right)$.
	\end{remark}

\subsection{Linear spline models}
	
In general, the curve $\overline{\psi}(\alpha)$ is an infinite dimensional representation of the object shape, and so cannot implemented in a tracking system. This problem can be solved by approximating $\overline{\psi}(\alpha)$ as a closed polygonal chain with a finite number of vertices, that is a linear spline.
	
	Let $\overline{V}_1,\dots,\overline{V}_n \in \mathbb{R}^2$ be a sequence of $n > 2$ vertices in barycentric coordinates and let
	\begin{equation}
	\overline{S} \triangleq \left[\begin{array}{cccc}
	\overline{V}_1' & \cdots & \overline{V}_n'
	\end{array}\right]'
	 \end{equation}
	be its vector representation. 
	Then, the linear spline $\kappa\left(\alpha; \overline{S}\right)$ with control points in $\overline{V}_1,\dots, \overline{V}_n$ is defined as the piece-wise linear 
 %{\color{red} parametric}
 curve $\kappa :\alpha \in [0,1] \mapsto \mathbb{R}^2$ given by \cite{PiegTill96}
	\begin{align}
	\label{blending1}
	\kappa\left(\alpha; \overline{S}\right) &\triangleq 
	(\mathsf{B}(\alpha)\otimes I)\,\overline{V}
	\qquad \alpha \in [0,1]\\
	\label{wrapped}
	\overline{V}&\triangleq \left[\begin{array}{cc}
	\hfil\overline{S}' & \overline{V}_1'
	\end{array} \right]'
	\end{align}	
 where we have introduced the blending vector $\mathsf{B}(\alpha)$ as
	\begin{align}
	\label{blending2}
	\mathsf{B}(\alpha) &\triangleq 
	\left[\begin{array}{ccc}
	\mathsf{B}_{1,n+1}(\alpha) & \cdots & \mathsf{B}_{n+1,n+1}(\alpha)
	\end{array}\right] \\
	\mathsf{B}_{i,n}(\alpha) &\triangleq \mathrm{tri}((n-1)\,\alpha-i+1) \\
	\mathrm{tri}(\alpha)&\triangleq(1-|\alpha|)\,\mathds{1}_{[-1,1]}(\alpha). \label{blending5}
	\end{align}
	The symbols $\otimes$, $I$ and $\mathds{1}_{\mathcal{A}}(x)$ denote respectively the Kronecker product, the identity matrix (of suitable dimension) and the indicator function of set $\mathcal{A}$ defined as $\mathds{1}_{\mathcal{A}}(x)\triangleq 1$ if $x\in \mathcal{A}$ or $\mathds{1}_{\mathcal{A}}(x)\triangleq 0$ otherwise. 
It can easily be checked that, by \eqref{wrapped}-\eqref{blending5}, \eqref{blending1} expresses the fact that, for $\alpha \in \left[ \frac{i-1}{n},\frac{i}{n} \right]$, the point $\kappa(\alpha;\overline{S})$ moves from $V_i$ to $V_{i+1}$ along the segment 
$\ell_i$ of the object contour.
Also note that (\ref{blending1}) allows to represent the contour and the surface of an object with the sets of points
	\begin{equation}
	\overline{\ell} \triangleq \left\{
	\kappa\left(\alpha;\overline{S}\right)
	\right\}_{\alpha\in[0,1]} \qquad
	\mathring{\overline{\ell}} \triangleq \mathrm{int}\,\overline{\ell}
	\end{equation}
	which are completely specified by a finite number of parameters, i.e. the $2n$ components of $\overline{S}$. Accordingly, in the linear spline framework, estimating the shape of an extended object is equivalent to jointly estimating the number $n$ and the positions $\overline{V}_1,\dots,\overline{V}_{n}$ of the vertices stored in $\overline{S}$. Since $n$ is a degree of freedom, (\ref{blending1}) can be qualified as a non-parametric shape model.

Since the linear spline curve $\kappa(\alpha;\overline{S})$ represents the shape of an extended object, the vector $\overline{S}$ cannot be chosen in a completely arbitrary way. 
According to the previous sections, three basic properties must be satisfied:
	\begin{itemize}
	\item the barycenter $\overline{g}$ falls in the origin;
	\item the left and right side contours $\{\kappa\left(\alpha;\overline{S}\right)\}_{\alpha \in [0,1/2)}$, 
	$\{\kappa\left(\alpha;\overline{S}\right)\}_{\alpha \in [1/2, 1]}$
	 are one the vertical reflection of the other, i.e. for all $\alpha \in (0,1/2)$ it holds that
	\begin{equation}
	\kappa\left(\alpha; \overline{S}\right)=\left[\begin{array}{cc}
	1 & \phantom{+}0 \\
	0 & -1
	\end{array}\right]\kappa\left(1-\alpha; \overline{S}\right);
	\end{equation}
	\item $\kappa\left(\alpha;\overline{S}\right)$ does not intersect itself, i.e. for all possible choices of $\alpha_1,\alpha_2\in[0,1]$ it holds that
	\begin{equation}
	\alpha_1\neq \alpha_2\\
	\,\,\Rightarrow\,\, \kappa\left(\alpha_1;\overline{S}\right) \neq 
	\kappa\left(\alpha_2;\overline{S}\right).
	\end{equation}
	\end{itemize} 
	In addition to the above properties, 
 it holds, without any loss of generality, that:
	\begin{itemize}
	\item $\overline{V}_1,\dots,\overline{V}_n$ are all distinct and every triplet of consecutive vertices (considering the wrapping convention $\overline{V}_{n+1}\triangleq \overline{V}_1, \overline{V}_{n+2}\triangleq \overline{V}_2$) is not collinear.
	\end{itemize}
	If $\overline{S}$ satisfies the above properties, then it is considered valid to represent the shape of an extended object and, in such a case, is referred to as a \textit{shape vector}.
	
	\begin{remark}
	Given a shape vector $\overline{V}$, the barycenter $\overline{g}$ of the linear spline $\kappa\left(\alpha; \overline{V}\right)$ can be computed via the \textup{shoelace formula \cite{shoelace}}
	\begin{equation}
	\label{eq:g}
	\overline{g} = \frac{1}{\mathsf{A}}\sum_{i=1}^n \underbrace{\left(\frac{\mathrm{det}\left[\begin{array}{cc}
	\overline{V}_i & \overline{V}_{i+1}
	\end{array}\right]}{2}\right)}_{\triangleq \mathsf{A}_i}\,\underbrace{\left(\frac{\overline{V}_i+\overline{V}_{i+1}}{3}\right)}_{\triangleq \overline{g}_i}
	\end{equation}
	where the wrapping convention $\overline{V}_{n+1}\triangleq \overline{V}_1$ is adopted. Note that $\overline{g}_i$ and $\mathsf{A}_i$ are respectively the barycenter and the (algebraic) area of the triangle with two vertices in $\overline{V}_i, \overline{V}_{i+1}$ and one vertex in the origin. Consequently,
	\begin{equation}
	\label{eq:A}
	\mathsf{A}\triangleq\sum_{i=1}^n \mathsf{A}_i
	\end{equation}
	represents the area encircled by the spline $\kappa\left(\alpha; \overline{V}\right)$.
	\end{remark}
	
	Thanks to the properties of the spline curves, the expression in world coordinates of the contour $\overline{\ell}$ can be obtained in two steps \cite{PiegTill96}. Given a desired location $g$ and heading $h$, the first step is to compute the shape vector in world coordinates, i.e.
	\begin{align}
	\label{eq:loc2wrld1}
	S&\triangleq \left[\begin{array}{ccc}
	V_1' & \cdots & V_n'
	\end{array}\right]'\\
	\label{eq:loc2wrld2}
	V_i &\triangleq U(h)\,\overline{V}_i + g  \qquad i = 1,\dots,n.
	\end{align}
	The second step is to interpolate the shape vector $S$, finally obtaining
	\begin{equation}
	\ell \triangleq \left\{
	\kappa\left(\alpha; S\right)
	\right\}_{\alpha\in[0,1]} \qquad
	\mathring{\ell} \triangleq \mathrm{int}\,\,\ell.
	\end{equation}
	
	Equations (\ref{eq:loc2wrld1})-(\ref{eq:loc2wrld2}) can be equivalently expressed in compact form in terms of the \textit{shape dewhitener}, which is defined by the affine transformation
	\begin{align}
	\label{direct}
	S  &= \mathsf{U}(h)\,\overline{S} + \mathsf{g}(g)\\
	\mathsf{U}(h)&\triangleq I \otimes U(h) \\
	\mathsf{g}(g)&\triangleq \big[\begin{array}{ccc}
	1 & \cdots & 1
	\end{array}\big]' \otimes \,g.
	\end{align}

\subsection{Linear spline partitioning}
	
	Let the contour curve $\psi(\alpha)$ be a linear spline $\kappa(\alpha; S)$ for some shape vector $S$ composed by a sequence of $n$ vertices $V_1,\dots,V_n$. Then, the domains $\ell$ (polygonal contour) and $\mathring{\ell}$ (polygonal surface) can be partitioned in elementary subdomains $\{\ell_i\}_{i=1}^{n}$ and $\{\mathring{\ell}_i\}_{i=1}^{n-2}$, that is
	\begin{align}
	\label{eq:partition1}
	&\ell=\bigcup_{i=1}^{n} \ell_i\\
	&\ell_i\cap \ell_{j\neq i} =\varnothing.
	\end{align}
	and
	\begin{align}
	\label{eq:partition2}
	&\mathring{\ell}=\bigcup_{i=1}^{n-2} \mathring{\ell}_i\\
	&\mathring{\ell}_i\cap \mathring{\ell}_{j\neq i} =\varnothing.
	\end{align}
	Indeed, in the contour case a generic polygonal chain $\ell$ with $n$ vertices can be split into its $n$ edges $\{\ell_i\}_{i=1}^n$,
	so that (\ref{eq:partition1}) takes the specific form
	\begin{align}
	\ell   &=\bigcup_{i=1}^n  \underbrace{\left\{\kappa\left(\alpha; \left[V_i'\,\,V_{i+1}'\right]'\right)\right\}_{\alpha\in[0,1]}}_{\triangleq \ell_i}.
	\end{align}
	where
	\begin{equation}
	\kappa\left(\alpha; \left[V_i'\,\,V_{i+1}'\right]'\right) \triangleq (1-\alpha)\,V_i+\alpha\,V_{i+1} \qquad \alpha\in[0,1].
	\end{equation}
	In the surface case, a generic polygonal surface $\mathring{\ell}$ can be split into $n-2$ triangles $\{\mathring{\ell}_i\}_{i=1}^{n-2}$, so that (\ref{eq:partition2}) takes the specific form
	\begin{align}
	\mathring{\ell}   &=\bigcup_{i=1}^{n-2}  \underbrace{ \mathrm{int}\left\{\kappa\left(\alpha; \left[V_i^1{}'\,\,V_{i}^{2}{}'\,\,V_{i}^3{}'\right]'\right)\right\}_{\alpha\in[0,1]}}_{\triangleq \mathring{\ell}_i}
	\end{align}
	where $V_k^j \in\{V_i\}_{i=1}^n$ for all $k\in\{1,\dots,n-2\}$ and $j\in\{1,2,3\}$. It is worth to point out that, as long as all vertices are distinct and no collinear triplets of consecutive vertices exist, the triangulation $\mathcal{T}\triangleq\{(V_i^1, V_i^2, V_i^3)\}_{i=1}^{n-2}$ can be generated, for example, by employing the \textit{ear clipping} method \cite{earclip}.
	
Partitions (\ref{eq:partition1}), (\ref{eq:partition2}) allow to write the uniform densities $\mathcal{U}(z; \ell)$, $\mathcal{U}(z; \mathring{\ell})$ in a particularly convenient form  that is a mixture of uniform densities over the elementary subdomains $\{\ell_i\}_{i=1}^{n}$ and $\{\mathring{\ell}_i\}_{i=1}^{n-2}$.

Consider the contour case, so that
\begin{equation}
\mathcal{U}(z;\ell)=\frac{\mathds{1}_{\ell}(z)}{|\ell|}
\end{equation} 
 due to the additivity of the characteristic function, the following decomposition holds
	\begin{equation}
	\label{eq:decomp}
	\mathds{1}_{\ell}(z)=\sum_{i=1}^{n}\mathds{1}_{\ell_i}(z)
	\end{equation}
	allowing to write 
	\begin{align}
		\label{eq:uniformMix}
	\mathcal{U}(z; \ell) &=\sum_{i=1}^{n} w_i\, \mathcal{U}(z; \ell_i)\\
	w_i &\triangleq \frac{|\ell_i|}{\sum_{j=1}^{n} |\ell_j|} \quad i=1,\dots,n.
	\end{align}
Note that to get the expression of the weight, we used the fact that
\begin{equation}
|\ell|=\sum_{j=1}^{n} |\ell_j|
\end{equation}
which, once again, is a consequence of (\ref{eq:partition1}). Clearly, the length of the elementary segment $\ell_i$ is given by
\begin{equation}
\label{eq:contweight}
|\ell_i| = \lVert V_{i+1}-V_i\rVert.
\end{equation}

Following the same rational, it turns out for the surface case that
\begin{align}
		\label{eq:uniformMix2}
	\mathcal{U}(z; \mathring{\ell}) &=\sum_{i=1}^{n-2} w_i\, \mathcal{U}(z; \mathring{\ell}_i)\\
	w_i &\triangleq \frac{|\mathring{\ell}_i|}{\sum_{j=1}^{n} |\mathring{\ell}_j|} \quad i=1,\dots,n.
	\end{align}

According to the shoelace formula, the area of the elementary triangle $\mathring{\ell}_i$ is given by
\begin{equation}
|\mathring{\ell}_i| = \sum_{j=1}^3 \frac{\mathrm{det}[V_i^j\,\,V_i^{j+1}]}{2}
\end{equation}
where, likewise in the previous section, the wrapping convention $V_i^{4}\triangleq V_i^{1}$ is considered. 

\subsection{Exact likelihood}
	A closed form expression for the single measurement contour likelihood $\mathcal{\ell}(y|c)$
	is available whenever the scattering domain $c$ is the contour $\ell$ of a polygonal chain.
	To obtain such expression, let us first consider the special problem in which the scattering domain is just a single and generic edge $\ell_i, i\in\{1,\dots,n\}$. 

In terms of the parametrization 
	\begin{equation}
	z(\alpha)\triangleq \kappa\left(\alpha; \left[V_i'\,\,V_{i+1}'\right]'\right) \qquad \alpha\in[0,1]
	\end{equation}
the relative likelihood can be  written as
	\begin{equation}
	\begin{aligned}
	\mathcal{L}(y|\ell_i)&=\int \mathcal{N}(y-z(\alpha);0, R )\,\,\mathcal{U}(z(\alpha); \ell_i)\,\,\underbrace{\lVert \dot{z}(\alpha)\rVert}_{=|\ell_i|} \,\,\mathrm{d}\alpha \\
	&=\int \mathcal{N}(y-z(\alpha);0, R)\,\,\underbrace{\mathds{1}_{\ell_i}(z(\alpha))}_{=\mathds{1}_{[0,1]}(\alpha)}\,\,\mathrm{d}\alpha\\
	&=\int_0^1 \mathcal{N}(A_i\alpha+B_i; 0, R)\,\,\mathrm{d}\alpha
	\end{aligned}
	\end{equation}
	where $A_i$ and $B_i$ are given by 
	\begin{align}
	A_i &\triangleq V_i-V_{i+1}\\
	B_i &\triangleq y-V_i
	\end{align}
	and %it was noted that 
	\begin{equation}
	\mathcal{U}(z;\ell_i) = \frac{\mathds{1}_{\ell_i}(z)}{|\ell_i|}.
	\end{equation}
	 Now, taking advantage of the square completion
	 \begin{equation}
	\label{eq:squareComp}
	\begin{aligned}
	\lVert A_i\alpha+B_i \rVert_{R^{-1}}^2 &\triangleq (A_i\alpha+B_i)' R^{-1} (A_i\alpha+B_i)\\
	&= \frac{1}{(A_i\!{}' R^{-1} A_i)^{-1}}\left(\alpha+\frac{B_i\!{}'R^{-1}A_i}{A_i\!{}' R^{-1}A_i}\right)^2 \\
	&\qquad+B_i\!{}'R^{-1}B_i-\left(\frac{B_i\!{}'R^{-1}A_i}{\sqrt{A_i\!{}' R^{-1}A_i}}\right)^2,
	\end{aligned}
	\end{equation}
	 the integrand can be factorized as %follows

	\begin{equation}
	\label{eq:GaussFact}
	\begin{aligned}
	\mathcal{N}(A_i\alpha+B_i;0,R)&=\frac{\mathcal{N}(B_i;0,R)}{\mathcal{N}\left(\frac{B_i\!{}'R^{-1}A_i}{\sqrt{A_i\!{}'R^{-1}A_i}};0, 1\right)}\\
	&\quad\times\frac{\mathcal{N}\left(\alpha;-\frac{B_i\!{}'R^{-1}A_i}{A_i\!{}'R^{-1}A_i}, \frac{1}{A_i\!{}'R^{-1}A_i}\right)}{\sqrt{A_i\!{}' R^{-1}A_i}}.
	\end{aligned}
	\end{equation}

	Let $\Phi:\mathbb{R}\mapsto[0,1]$ be the univariate standard cumulative distribution function, i.e.
	\begin{equation}
	\Phi(x)\triangleq \frac{1}{\sqrt{2\pi}}\int_{-\infty}^{x} \exp\left(-\frac{\beta^2}{2}\right)\,\,\mathrm{d}\beta.\end{equation}
	Then, the single edge likelihood gets the closed form
	\begin{equation}
	\label{eq:exactlikyci}
	\mathcal{L}(y|\ell_i)=\frac{\mathcal{N}(B_i;0,R)}{\mathcal{N}\left(\frac{B_i\!{}'R^{-1}A_i}{\sqrt{A_i\!{}'R^{-1}A_i}};0, 1\right)}\frac{\Delta \Phi(A_i, B_i; R)}{\sqrt{A_i' R^{-1} A_i}}
	\end{equation}
	\begin{equation}
	\begin{aligned}
	\Delta \Phi(A_i, B_i; R)&\triangleq
	\Phi\left(\frac{A_i\!{}' R^{-1} A_i+B_i\!{}'R^{-1}A_i}{\sqrt{A_i\!{}' R^{-1} A_i}}\right)\\
	&\qquad-\Phi\left(\frac{B_i\!{}'R^{-1}A_i}{\sqrt{A_i\!{}' R^{-1} A_i}}\right).
	\end{aligned}
	\end{equation}
	
	\begin{remark}As sanity check, note that if the edge $\ell_i$ collapses into a single point $V_i$ then, as expected, (\ref{eq:exactlikyci}) reduces to a Gaussian density centered in $V_i$, with covariance $R$, evaluated at $y$ (that is, $\mathcal{N}(B_i;0, R)$). 
 To see why this is true, let $V_{i+1}\triangleq V_i+\rho\, \left[\cos \theta\,\,\sin \theta\right]'$. Then, for suitable functions $\phi_1(\theta)$ and $\phi_2(\theta)$, one can write $B_i\!{}'R^{-1}A_i=\rho\,\phi_1(\theta)$ and $A_i\!{}' R^{-1}A_i=\rho^2\,\phi_2(\theta)$. Now, regardless the specific value of $\theta$, according to (\ref{eq:exactlikyci}) it holds that
	\begin{equation}
	\begin{aligned}
	&\lim_{\rho \to 0} \frac{\mathcal{L}(y|\ell_i)}{\mathcal{N}(B_i;0,R)}\bigg|_{\substack{B_i\!{}'R^{-1}A_i =\rho\,\,\phi_1(\theta)\\A_i\!{}'R^{-1}A_i=\rho^2\phi_2(\theta)}}=\\
	&\qquad\qquad\frac{1}{\mathcal{N}\left(\frac{\phi_1(\theta)}{\sqrt{\phi_2(\theta)}};0, 1\right)}\,\frac{\mathrm{d}\Phi(x)}{\mathrm{d}x}\Bigg|_{x=\frac{\phi_2(\theta)}{\sqrt{\phi_1(\theta)}}}=1
	\end{aligned}
	\end{equation}
	which proves the above claim.
	\end{remark}
	
	Now consider the general case where the scattering domain is a polygonal chain $\ell$ consisting of $n$ edges $\{\ell_i\}_{i=1}^n$. According to (\ref{eq:exactlikyci}) and (\ref{eq:uniformMix}), 
	\begin{equation}
	\label{eq:exactliky}
	\begin{aligned}
	\mathcal{L}(y|\ell)
	=\sum_{i=1}^n w_i\,\mathcal{L}(y|\ell_i)
	\end{aligned}
	\end{equation}
	which is nothing but a mixture of single measurement likelihoods over the individual edges $\ell_i$. In particular, due to (\ref{eq:contweight}), the weights of the mixture take the following specific form
	\begin{equation}
	w_i\triangleq \frac{\lVert V_{i+1}-V_i \rVert}{\sum_{j=1}^n \lVert V_{j+1}-V_j \rVert} \qquad i=1,\dots,n
	\end{equation}
	where, like in the previous sections, $V_{n+1}\triangleq V_1$.
	Finally, substituting (\ref{eq:exactliky}) in (\ref{eq:exactDatasetlik}) gives the exact dataset likelihood for the contour case, that is
	\begin{equation}
	\label{eq:exactlikY}
	\begin{aligned}
	\mathcal{L}(Y|\ell) &= 
	\mathcal{B}\left(m;\ell\right)\,\prod_{j=1}^m \mathcal{L}\left(y^{(j)}|\ell\right)
	\end{aligned}
	\end{equation}
	where, according to (\ref{eq:exactDatasetlik}), $\prod_{j=1}^m \mathcal{L}\triangleq 1$ if $m=0$. 
	
	Notice that the computational cost of (\ref{eq:exactlikY}) is $\mathcal{O}(mn)$ where $m$ is the number of measurements and $n$ the number of vertices of the linear spline.

\subsection{Monte Carlo likelihood}
	
	Unfortunately, finding a closed-form expression for $\mathcal{L}(Y|\mathring{\ell})$ is a hard, if not impossible, task. On the other hand, allowing for a certain approximation error, the integral can be easily computed with any numerical integration technique. Start by considering the single measurement likelihood $\mathcal{L}(y|\mathring{\ell})$. A natural solution to compute $\mathcal{L}(y|\mathring{\ell})$ is suggested by the Monte Carlo method, leading to the approximate likelihood
	\begin{equation}
	\label{eq:MCylik}
	\begin{aligned}
	\mathcal{L}^{\text{MC}}(y|\mathring{\ell})\triangleq \frac{1}{N} \sum_{i=1}^N \mathcal{N}\left(y;z_{\mathring{\ell}}^{(i)}, R\right)
	\end{aligned}
	\end{equation}
	where the empirical particle set $Z_{\mathring{\ell}}\triangleq\{z_{\mathring{\ell}}^{(i)}\}_{i=1}^N$ is obtained by uniformly sampling the domain $\mathring{\ell}$, i.e., 
	\begin{equation}
	z_{\mathring{\ell}}^{(i)}\sim \mathcal{U}(z; \mathring{\ell})\qquad i=1,2,\dots,N.
	\end{equation}
	Note that the number $N\in\mathbb{N}$ of particles is a degree of freedom to trade off accuracy versus computational cost of the approximation (both increasing with $N$).
	Combining (\ref{eq:MCylik}) with (\ref{eq:exactDatasetlik}) yields the %definition of 
 Monte Carlo dataset likelihood
	\begin{equation}
	\label{eq:MCDatasetlik}
	\mathcal{L}^{\text{MC}}(Y|\mathring{\ell})\triangleq 
	\mathcal{B}\left(m;\mathring{\ell}\right) \,
	\prod_{j=1}^m \mathcal{L}^{\text{MC}}\left(y^{(j)}|\mathring{\ell}\right)
	\end{equation}
	where, according to (\ref{eq:exactDatasetlik}), it is assumed  by convention that $\prod_{j=1}^m \mathcal{L}^{\text{MC}}\triangleq 1$ when $m=0$.
	
	Note that, given the particle set  $Z_{\mathring{\ell}}$, the cost for the Monte Carlo likelihood computation (\ref{eq:MCDatasetlik}) is $\mathcal{O}(mN)$, since each dataset measurement $y^{(j)}$ has to be compared with each particle $z_{\mathring{\ell}}^{(i)}$. 
	
	At this stage, the problem of computing the dataset likelihood is reduced to the simpler one of developing a uniform sampler for the generic scattering domain $\mathring{\ell}$.

	\begin{figure}
	\centering
	\includegraphics[scale=0.4]{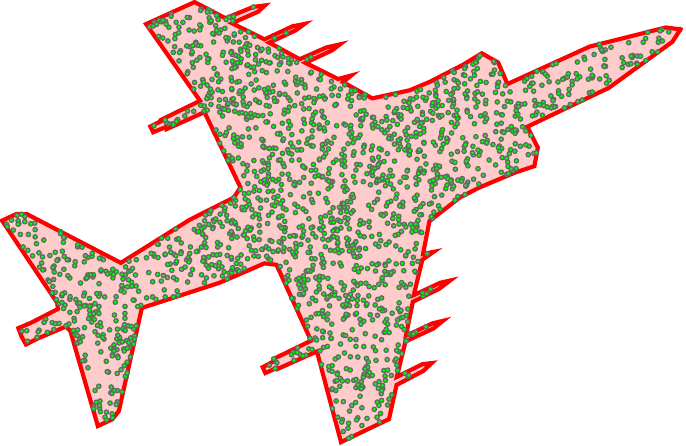}
	\caption{visualization of $2\cdot10^3$ points uniformly distributed over the object surface.}
	\end{figure}
	
	Equation (\ref{eq:uniformMix2}) suggests a possible strategy to draw a point $z$ uniformly distributed over $\mathring{\ell}$: (step 1) generate a random integer $i$ according to the discrete distribution $\{w_i\}_{i=1}^{n-2}$; (step 2) generate a particle $z$ according to the uniform distribution $\mathcal{U}(z;\mathring{\ell}_i)$. Since step 1 can be carried out by employing the \textit{inverse transform sampling} method \cite{devroye2006nonuniform}, the main problem is to find a method to sample the elementary subdomain $\mathring{\ell}_i$.
To this end,
	it has been shown in \cite{osada2002shape} that the baricentric combination
	\begin{align} 
	&z\triangleq (1-\sqrt{\alpha_1})V_i^1+\sqrt{\alpha_1}(1-\alpha_2)V_i^2+\sqrt{\alpha_1}\alpha_2 V_i^3 \\
	&\alpha_1,\alpha_2\sim \mathcal{U}(\alpha; [0,1])
	\end{align}
	generates $z\sim\mathcal{U}(z;\mathring{\ell}_i)$.
	
	Exploiting the inverse transform method, sampling $N$ particles from a shape with $n$ vertices involves $\mathcal{O}(nN)$ cost. As a consequence, the overall computational complexity of the Monte Carlo likelihood computation is $\mathcal{O}((m+n)\,N)$.
	Note that, for the surface case, the cost of the triangulation of the polygonal surface $\mathring{\ell}$ is omitted due to the fact that such task can be performed once offline\footnote{For the sake of completeness, the ear clipping method has $\mathcal{O}(n^3)$ cost. Consequently, the computational cost of the sampler is no longer linear in $n$ if the required triangulation is not given.}.

\begin{remark}Given a shape vector $S$ with $n$ vertices, any decimation algorithm, e.g. the \textup{RDP method \cite{Douglas1973RDP}}, can be applied to $S$ to obtain an approximated shape $\tilde{S}$ having $\tilde{n}<n$ vertices. Consequently, the computational cost of the %{\color{red} footpoint or ???} 
exact likelihood can be reduced by replacing $S$ with its approximation $\tilde{S}$.
	\end{remark}

\section{Estimation}
	
\subsection{Tracker}
	
Following \cite{tesori2023lomem}, it is assumed that the object longitudinal axis and the object velocity are aligned, a condition that is usually satisfied in many applications of interest. A class of object motion models
that conveniently take advantage of this assumption  is the so-called class of $\lambda:o$ models \cite{tesori2022lambda}, 
that include, in the object state, derivatives of the speed and turning rate up to orders $\lambda-1$ and $o-1$ respectively.
Specifically, we adopt the $2:1$ model with state vector
\begin{equation}
	x \triangleq \left[\begin{array}{ccccc}
	g' & h & s & \dot{s} & \omega
	\end{array}\right]' \in \mathbb R^6
	\end{equation}
 including barycenter position $g \in \mathbb R^2$, heading $h$,
 speed $s$ (time derivative of the velocity vector modulus $\lVert\dot{g}\rVert$), speed rate $\dot s$, turning rate $\omega$ (time derivative of the heading angle $h$).
 Considering also the process noise
	\begin{equation}
	w\triangleq\left[\begin{array}{ccccc}
	{w^g}' & w^h & w^s & w^{\dot{s}} & w^{\omega}
	\end{array}
	\right]' \sim \mathcal{N}(w;0,Q)
	\end{equation}
and exploiting Tustin discretization with sampling (scan) period $T$, the model is described
by the following discrete-time state equations in the scan index $k$
	\begin{align}
	\label{pk+1}
	g_{k+1} &= g_k + T\,f(x_k,w_k) +w_k^g \\
	\label{hk+1}
	h_{k+1} &= h_k + T \omega_k+w_k^{\omega} \\
	\label{sk+1}
	s_{k+1} &=s_k+T\dot{s}_k+w_k^s \\
	\dot{s}_{k+1} &= \dot{s}_k+w_k^{\dot{s}} \\
	\label{omegak+1}
	\omega_{k+1} &=\omega_k+w_k^{\omega}
	\end{align}
	which turn out to be nonlinear due to the Tustin step
	\begin{equation}
 \label{Tustin}
	f(x_k,w_k) \triangleq 
	\frac{1}{2}\left(s_{k+1}\left[\begin{array}{c}
	\cos h_{k+1} \\
	\sin h_{k+1}
	\end{array}\right] + 
	s_{k}\left[\begin{array}{c}
	\cos h_{k} \\
	\sin h_{k}
	\end{array}\right] 
	\right) .
	\end{equation}
	%where $h_{k+1}$, $s_{k+1}$ are respectively given by (\ref{hk+1})  and (\ref{sk+1}).}
	This motion model is tailored for maneuvering objects since it allows to track trajectories characterized by time-varying speed and turning rate.
	Equations (\ref{pk+1})-(\ref{Tustin}) can be rewritten in compact form as
	\begin{align}
	\label{pred}
	x_{k+1}&=x_k+T \cdot F\left(x_k,w_k\right)+ w_k \\
	\label{eq:F}
	F\left(x,w\right) &\triangleq
	\left[\begin{array}{cccc}
	f'(x,w) &
	\omega &
	\dot{s} &
	0
	\end{array}\right]'.
	\end{align}
	Based on (\ref{pred}), any nonlinear technique can be exploited for prediction. Since (\ref{pred}) can be differentiated analytically, a simple choice is to adopt the extended Kalman filter (but the unscented Kalman filter could be used as well). 
	
	Let $x_{k-1|k-1}$ and $P_{k-1|k-1}$ be the corrected estimate and covariance generated at the previous sensor scan. Then, the proposed extended Kalman predictor takes the following form
	\begin{align}
	\label{eq:pred1}
	x_{k|k-1}&\triangleq x_{k-1|k-1}+T \, F\left(x_{k-1|k-1},0\right) \\
	\label{eq:pred2}
	P_{k|k-1}&\triangleq J_x P_{k-1|k-1} J_x' + J_w Q J_w' \\
	\label{eq:J}
	J_x &\triangleq I+T\,\frac{\partial F}{\partial x}\bigg|_{\substack{x=x_{k-1|k-1}\\ w=0\phantom{++++}\,\,\,}}\\
	J_w &\triangleq T\,\frac{\partial F}{\partial w}\bigg|_{\substack{x=x_{k-1|k-1}\\ w=0\phantom{++++}\,\,\,}}.
	\end{align}
%The analytic expressions of the Jacobian $\frac{\partial F}{\partial x}$ are given in the appendix.

	Following \cite{tesori2023lomem}, the proposed corrector performs a pre-processing transformation in which the dataset is mapped onto a static estimate $\hat{g}_k$ of the object position $g_k$ and a static estimate $\hat{h}_k$ of the object heading $h_k$. Then, a standard Kalman corrector is applied to generate the corrected kinematic state estimate. Assume for the moment to have at hand the two current static estimates $\hat{g}_k, \hat{\theta}_k$. The equivalent measurement model between the virtual measurement $y_k^{\text{v}}\triangleq [\hat{g}_k'\,\,\hat{\theta}_k]'$ and the kinematic state $x_k$ is given by
	\begin{align}
	y_k^{\text{v}}&\triangleq C x_k + e_k  \\
	e_k &\sim \mathcal{N}(\cdot;0, E) \\
	C&\triangleq \left[\begin{array}{cc}
	I & 0
	\end{array}\right].
	\end{align}
	Note that $e_k$ is not the original measurement noise $v_k$ affecting the measurements, but rather it represents the estimation error of the static estimates $\hat{g}_k, \hat{\theta}_k$. For the sake of simplicity, such error is considered as a Gaussian variable with zero-mean and covariance $E$ (treated as a tuning parameter).
	
	Let $x_{k|k-1}$ and $P_{k|k-1}$ be the predicted estimate and covariance generated at the previous sensor scan. Then, the proposed corrector gets the following form
	\begin{align}
	\label{eq:corr1}
	x_{k|k} &\triangleq (I-L_k C)\,x_{k|k-1}+L_k\, y_k \\
	\label{eq:corr2}
	P_{k|k} &\triangleq (I-L_k C) P_{k|k-1}\\
	\label{eq:corr3}
	L_k &\triangleq P_{k|k-1} C' \left(C P_{k|k-1}C'+E\right)^{-1}.
	\end{align}

	Now consider the problem of computing the static estimates $\hat{g}_k$ and $\hat{h}_k$ given the current dataset $Y_k$. According to the proposed measurement model, the measurements contained in the dataset are uniformly scattered around the object barycenter (either in the contour case or surface case). This fact suggests to estimate the object position with the dataset barycenter, i.e. the mean measurement
	\begin{equation}
	\label{eq:hatpk}
	\hat{g}_k \triangleq \frac{1}{m_k} \sum_{j=1}^{m_k} y_k^{(j)}.
	\end{equation}
	For what concerns the heading estimation, since the velocity vector is assumed to be aligned to the longitudinal axis, a possible static estimate is
	\begin{equation}
	\label{eq:hatthetak}
	\hat{h}_k \triangleq \angle\left(\frac{\hat{g}_k-g_{k-1|k-1}}{\lVert \hat{g}_k-g_{k-1|k-1} \rVert}\right)
	\end{equation}
	where $g_{k-1|k-1}$ is the corrected barycenter at the previous scan (i.e. the first two components of $x_{k-1|k-1}$) and $\angle \zeta$ is the angle that the generic bivariate vector $\zeta$ forms with respect to the horizontal axis of the world reference system. Note that (\ref{eq:hatthetak}) requires that the object speed is non-null, a condition that holds in air surveillance applications involving airplanes without hovering capabilities.
	
	\begin{algorithm}[tb]
	\SetAlgoLined
	\vspace{0.2cm}
	\KwIn{$Y_k, x_{k-1|k-1}, P_{k-1|k-1}$}
	\KwOut{$x_{k|k}, P_{k|k}$}
	\KwPar{$Q, E$}
	\hrulefill
	\vspace{0.1cm}\\
	\textbf{1)} evaluate the Jacobian (\ref{eq:J}); \\
	\textbf{2)} compute the prediction (\ref{eq:pred1}), (\ref{eq:pred2}); \\
	\textbf{3)} compute the static estimates (\ref{eq:hatpk}), (\ref{eq:hatthetak}); \\
	\textbf{4)} compute the correction gain (\ref{eq:corr3});\\
	\textbf{5)} compute the correction (\ref{eq:corr1}), (\ref{eq:corr2}). 
	\caption{\textbf{Tracker}.}
	\label{alg:tracker}
	\end{algorithm}

	\begin{algorithm}[tb]
	\SetAlgoLined
	\vspace{0.2cm}
	\KwIn{$Y_k, x_{k|k}, \{p_{k-1}(\iota)\}_{\iota=1}^{\mathcal{I}}$}
	\KwOut{$\{p_{k}(\iota)\}_{\iota=1}^{\mathcal{I}}, \hat{S}_k$}
	\KwPar{$\overline{\mathcal{S}},\Delta R, \delta, (\text{eventually } N)$}
	\hrulefill
	\vspace{0.1cm}\\
	\textbf{1)} extract ${g}_{k|k}$, ${h}_{k|k}$ from $x_{k|k}$; \\
	\textbf{2)} define $\overline{Y}$ via the dataset whitener 
	\begin{equation*}
	\overline{y}^{(j)} \triangleq U\left({h}_{k|k}\right)'\left(y_k^{(j)}-{g}_{k|k}\right) \qquad j=1,\dots,m;
	\end{equation*}\\
	\textbf{3)} based on $R_{+}$, perform the parallel update in terms of (\ref{eq:MCDatasetlik}), or (\ref{eq:exactlikY})\\[0.1cm]
	\For{$\!\!\parallel\,\iota=1,\dots,\mathcal{I}$}{
	$\tilde{p}_{k}(\iota)\triangleq \mathcal{L}\left(\overline{Y}|\iota\right)\,\left[\frac{\mathcal{I}\delta-1}{\mathcal{I}-1}p_{k-1}(\iota)+\frac{1-\delta}{\mathcal{I}-1}\right];$
	}
	$p_{k}(\iota)\triangleq \tilde{p}_k(\iota) / \sum_{\nu=1}^{\mathcal{I}} \tilde{p}_k(\nu) \qquad \iota =1,\dots, \mathcal{I};$\\[0.1cm]
	\textbf{4)} find the modal class
	\begin{equation*}
	\iota^* \triangleq \arg \max_{\iota\in\{1,\dots,\mathcal{I}\}} p_k(\iota);
	\end{equation*}\\
	\textbf{5)} define $\hat{S}_k$ via the shape dewhitener
	\begin{equation*}
	\hat{S}_k \triangleq \mathsf{U}\left(h_{k|k}\right)\overline{S}^{(\iota^*)}+\mathsf{g}(g_{k|k}).
	\end{equation*}
	\caption{\textbf{Shaper}.}
	\label{alg:shaper}
	\end{algorithm}

\subsection{Dictionary} 
	
	The proposed algorithm that jointly estimates object position, orientation and shape is referred to as \textit{shaper}. Essentially, the shaper is a Bayesian classifier that selects the estimated shape among a finite number $\mathcal{I}>1$ of different primitive shapes. Such primitive shapes are supposed to be known and stored in the set
	\begin{equation}
	\label{eq:classes}
	\overline{\mathcal{S}} \triangleq \left\{\overline{S}^{(\iota)}\right\}_{\iota=1}^{\mathcal{I}}
	\end{equation}
	which is referred to as (shape) \textit{dictionary}. 
 %{\color{red} The dictionary provides a degree of freedom for the shaper ???.} 
 Each primitive shape is encoded by a shape vector expressed in barycentric coordinates, i.e.,
	\begin{equation}
	\overline{S}^{(\iota)}\triangleq \left[\begin{array}{ccc}
	\overline{V}_1^{(\iota)'} & \cdots &
	\overline{V}_{n^{(\iota)}}^{(\iota)'}
	\end{array}\right]'
	\qquad \iota=1,\dots,\mathcal{I}.
	\end{equation}
	Note that each shape vector can have a proper number $n^{(\iota)}$ of vertices. A possible way to quantify the amount of information stored in the dictionary is provided by the total amount of vertices, denoted as
	\begin{equation}
	\mathrm{com}\left(\overline{\mathcal{S}}\right)\triangleq \sum_{\iota=1}^{\mathcal{I}} n^{(\iota)}
	\end{equation}
	and referred to as \textit{dictionary complexity}. Introducing the dictionary \textit{average complexity} as the average number of vertices throughout the dictionary, that is
	\begin{equation}\overline{n}\triangleq \frac{1}{\mathcal{I}}\sum_{\iota=1}^{\mathcal{I}}n^{(\iota)},\end{equation}
	allows to conveniently identify the computational cost of the shaper in terms of the quadruple $(m,\overline{n},\mathcal{I}, N)$.
	
	Each shape vector encodes the scattering domains
	\begin{equation}
	\label{eq:contours}
	\begin{aligned}
\overline{\ell}^{(\iota)}&\triangleq\left\{\kappa\left(\alpha; \overline{S}^{(\iota)}\right)\right\}_{\alpha\in[0,1]}\\
	\mathring{\overline{\ell}}{}^{(\iota)}&\triangleq\text{int}\left\{\kappa\left(\alpha; \overline{S}^{(\iota)}\right)\right\}_{\alpha\in[0,1]}.
	\end{aligned}
	\end{equation}
	According to $(\ref{eq:classes})$, each shape vector identifies a specific object class $\iota$; according to $(\ref{eq:contours})$, each shape vector identifies a specific object contour $\overline{\ell}{}^{(\iota)}$. Thus, the shaper is an algorithm that treats the two problems of object classification and shape estimation as equivalent, solving them simultaneously using a single Bayesian classifier.

Beside shape vectors, the dictionary can store all information about an object that can be pre-computed before tracking. For example, an object class $\iota$ can be identified by the quadruple $\big(\overline{S}^{(\iota)}, |\mathring{\overline{\ell}}{}^{(\iota)}|, \mathcal{T}^{(\iota)}, \overline{\Gamma}^{(\iota)}\big)$
(i.e., object shape, size, triangulation and reflectivity)
.
	
	\subsection{Classifier}
	The goal of the classifier is to quantify the probability that a given shape vector generates the true object contour. 
	To achieve this objective, the classifier compares the current dataset $\overline{Y}_k$ expressed in local coordinates with respect to the primitive shapes stored in the dictionary, by employing the standard prediction-correction scheme.
	
Let $\{p_{k-1}(\iota)\}_{\iota=1}^{\mathcal{I}}$ be the prior probability distribution encoding the class belief up to the previous scan $k-1$.
Then, the predicted probability distribution at scan $k$ is defined according to the Chapman-Kolmogorov equation as
\begin{equation}
\label{eq:chap}
p_{k|k-1}(\iota) \triangleq \sum_{j=1}^\mathcal{I} \varphi\left(\iota |j\right)\,p_{k-1}(j) \qquad \iota=1,\dots,\mathcal{I}
\end{equation}
for some transition kernel $\varphi(\iota|j):\{1,\dots,\mathcal{I}\}^2\mapsto [0,1]$. A possible choice for $\varphi(\iota|j)$ is given by
\begin{equation}
\label{eq:phi}
\varphi(\iota|j)\triangleq \begin{cases}
\hfil \delta &\text{if } \iota=j \\
\hfil \frac{1-\delta}{\mathcal{I}-1} &\text{otherwise}\\
\end{cases}
\end{equation}
where $\delta\in[0,1]$ is a tuning parameter. Combining (\ref{eq:chap}), (\ref{eq:phi}) yields to the following simplified expression for the predicted distribution
	\begin{equation}
	\label{eq:chap2}
p_{k|k-1}(\iota) = \frac{\mathcal{I}\delta-1}{\mathcal{I}-1}\,p_{k-1}(\iota) +\frac{1-\delta}{\mathcal{I}-1}\qquad \iota=1,\dots,\mathcal{I}.
\end{equation}
A desirable property is that the predicted distribution preserves the class ranking of the prior distribution, in the sense that $p_{k-1}(i)>p_{k-1}(j) \Rightarrow p_{k|k-1}(i)>p_{k|k-1}(j)$.
As one can show, such property holds if and only if the tuning parameter satisfies $\delta>1/\mathcal{I}$. Consequently, a proper calibration can be obtained by choosing $\delta \in (1/\mathcal{I}, 1]$.

	 According to the Bayes rule, the predicted distribution is updated according to the dataset $\overline{Y}_k$ via 
	\begin{equation}
	\label{eq:post}
	p_{k}(\iota)\triangleq \frac{\mathcal{L}\left(\overline{Y}_k|\iota\right)\,p_{k|k-1}(\iota)}{\sum_{\nu=1}^{\mathcal{I}} \mathcal{L}\left(\overline{Y}_k|\nu\right)\,p_{k|k-1}(\nu)}
	\qquad \iota=1,\dots,\mathcal{I}
	\end{equation}
	where, in particular, the proposed method computes the likelihood values
	\begin{equation}
	\mathcal{L}\left(\overline{Y}_k|\iota\right)\triangleq \begin{cases}
	\hfil \mathcal{L}\left(\overline{Y}_k|\overline{\ell}^{(\iota)}\right) &
	\text{contour case}\\
	\hfil \mathcal{L}\left(\overline{Y}_k|\mathring{\overline{\ell}}{}^{(\iota)}\right) &
	\text{surface case}\\
	\end{cases}
	\end{equation}
 according to either (\ref{eq:exactlikY}) or
(\ref{eq:MCDatasetlik}) for the contour, or respectively, surface case. 
	
	In order to take into account the unavoidable uncertainty introduced by the dataset whitener, the generic dataset likelihood is evaluated in terms of the inflated covariance matrix
	\begin{equation}
	R_{+}\triangleq R + \Delta R
	\end{equation}
	where $\Delta R$ is an incremental covariance matrix that %, {\color{red} for the sake of simplicity ???}, 
 is considered as a tuning parameter.

	\subsection{Shaper}
	
	Consider a generic dataset $Y$ generated by an extended object located in an unknown position $g$ and oriented according to an unknown heading $h$. If an estimate $\hat{g}$ for the position $g$ and an estimate $\hat{h}$ for the heading $h$ are available, then a new dataset in local coordinates
	\begin{equation}
	\label{eq:localData}
	\overline{Y} \triangleq \left\{
	\overline{y}^{(j)}
	\right\}_{j=1}^{m} 
	\end{equation}
	can be computed by applying to each measurement $y\in Y$ the inverse roto-translation 
	\begin{equation}
	\label{inversedata1}
	\overline{y}^{(j)} \triangleq U\left(\hat{h}\right)' \left(y^{(j)}-\hat{g}\right) \qquad j=1,\dots, m.
	\end{equation}
	
	The new dataset $\overline{Y}$ is approximately centered in the origin and approximately aligned with the axes of the local reference system, and so can be compared with each shape vector stored in the dictionary. The map defined by (\ref{inversedata1}) is referred to as \textit{dataset whitener}.
	Based on the output of the classifier, the estimated shape is generated by selecting the shape vector (in world coordinates) relative to the modal class, i.e.
	\begin{align}
	%\hat{{\psi}}(\alpha)&\triangleq 
	%\kappa\left(\alpha; {S}^{(\iota^*)}\right) \qquad \alpha \in [0,1]\\
	{{S}^{(\iota^*)}}&\triangleq \mathsf{U}\left(\hat{h}\right)\, \overline{S}^{(\iota^*)}+\mathsf{g}\left(\hat{g}\right)\\
	\label{eq:modal}
	\iota^{*}&\triangleq \arg \max_{\iota\in\{1,\dots,\mathcal{I}\}} p_{k}(\iota).
	\end{align} 
	
	Depending on the considered likelihood model, the overall cost of the shaper takes the following order of magnitude:
	\begin{itemize}
	\item $\mathcal{O}((m+\overline{n})\,N\,\mathcal{I})$ if the dataset likelihood is computed using the Monte Carlo approximation;
	\item $\mathcal{O}(m\,\overline{n}\,\mathcal{I})$
	if the dataset likelihood is computed using the  exact formula.
	\end{itemize}
	
	The processing time can be reduced by observing that the value of each posterior probability $p_{k}(\iota)$ can be computed simultaneously with a parallel processor.

	\begin{figure}
	\centering
	\includegraphics[scale=0.3]{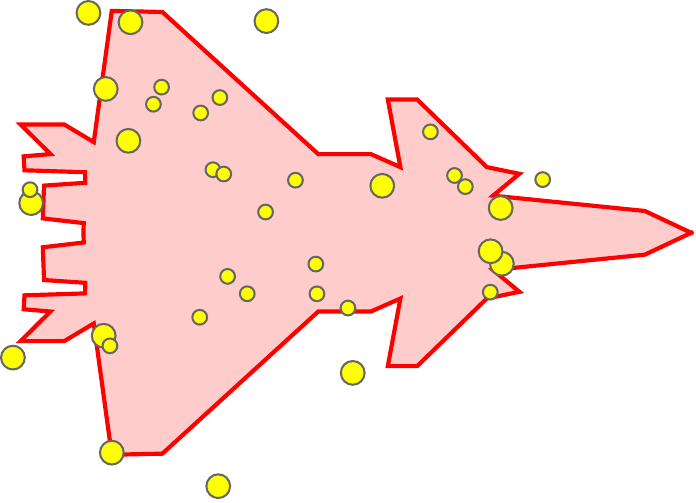}
	\caption{Sample scan of the classification test. 
	Contour (surface) measurements are shown as large (small) dots.
	}
	\end{figure}

\section{Numerical experiments}
	
	\subsection{Classification test}
	
	The classification test has the objective to provide an empirical evidence of the main characteristics of the likelihood models (\ref{eq:MCDatasetlik}), (\ref{eq:exactlikY}). Such models are respectively denoted as $\textbf{MC}$, and $\textbf{EC}$.

	An unknown and static object, which consists in a delta-wing fighter jet placed in the origin with null heading, is observed $100$ times by a contour and a surface sensor. The object radii are $\rho_{\max}\approx 13\,[\mathrm{m}]$ and $\rho_{\min}\approx 3\,[\mathrm{m}]$, while the contour length and area are $|\ell|\approx80\,[\mathrm{m}]$, 
	$|\mathring{\ell}|\approx120\,[\mathrm{m}^2]$.
	
	The sensors are affected by isotropic noise with standard deviation $\sigma \triangleq 1\,[\mathrm{m}]$. The sensors are characterized by the resolutions $\overline{r}\triangleq 5\,[\mathrm{m}]$, $\mathring{\overline{r}}\triangleq 5\,[\mathrm{m}^2]$ and lighting power $\overline{\eta}\triangleq 0.9$.

	At each scan, the likelihood models are evaluated considering a dictionary composed by $\mathcal{I}\triangleq 21$ primitive shapes relative to different aircraft. In particular, the considered dictionary includes the shapes of an airliner, several conventional fighter aircraft, several stealth fighter jets, two stealth bombers, several delta-wing aircraft, two propeller aircraft and more. The average complexity of the considered dictionary is $\lfloor\overline{n}\rceil= 55$.

	For simplicity, the reflectivity of each object is set to the ideal value $\overline{\Gamma}\triangleq 1$. The resulting expected dataset cardinalities are given by $\overline{m}_{\ell}\triangleq\overline{\mu}_{\ell}\,\overline{\pi}_{\ell}\approx 15$, $\overline{m}_{\mathring{\ell}}\triangleq\overline{\mu}_{\mathring{\ell}}\,\overline{\pi}_{\mathring{\ell}}\approx 22$. The Monte Carlo likelihood uses $N\triangleq 10^3$ particles.

	At each sensor scan, the maximum likelihood estimate (MLE) is computed for each likelihood model. If a MLE correctly identifies the shape of the unknown object then a flag is generated.
	The accuracy of a likelihood model is evaluated as the fraction of times that the relative MLE correctly identify the unknown object over the $100$ trials.
	
	In the contour case, $\textbf{EC}$ achieves an accuracy of $86\%$;
	in the surface case, $\textbf{MC}$ achieves an accuracy of $79\%$.

\subsection{Tracking test}

	The tracking test has the objective to asses the effectiveness of the proposed methods in solving extended object tracking problems. To achieve this goal, numerical experiments simulating an air surveillance problem are considered. The tracked object is a multi-role fighter jet, performing evasive maneuvers at high speed over the time span of one minute. 
	The fighter jet has radii $\rho_{\max}\approx 16\,\,[\mathrm{m}]$ and $\rho_{\min}\triangleq \min_{z\in\overline{c}}\lVert z \rVert\approx 3\,[\mathrm{m}]$, contour length $|\ell|\approx 95\,[\mathrm{m}]$ and contour area $|\mathring{\ell}|\approx 106\,[\mathrm{m}^2]$.
	
	Similarly to the classification test, tracking is carried out by simultaneously considering a contour and a surface sensor, resulting in the generation of two dataset at each scan.
	Both sensors are characterized by multiple resolutions $d(\overline{r})$, different levels $\sigma$ of inaccuracy (assuming isotropic noise), lighting power $\overline{\eta}\triangleq 0.9$ and sampling time $T\triangleq 0.1\,\left[\mathrm{s}\right]$.
	
	The object reflectivities are set to the ideal value $\overline{\Gamma}(d(\ell))\triangleq 1$. Moreover, the dictionary employed in this test is the same as the one used in the classification test.

	The global algorithm, FL:OREO, is compared with state of the art extended object estimators, namely the original RH estimator from \cite{baum2014RHM} and its GP variant from \cite{wahlstrom2015GP}. Based on the likelihood model considered, two versions of FL:OREO are considered: $\textbf{EC}_{\ell}$ (Exact Contour), $\textbf{MC}_{\mathring{\ell}}$ (Monte Carlo).
	
For each random hypersuperface estimator, two versions are considered: $\textbf{RH}_{\ell}$ (original RH, contour case), $\textbf{RH}_{\mathring{\ell}}$ (original RH, surface case), $\textbf{GP}_{\ell}$ (GP RH, contour case), $\textbf{GP}_{\mathring{\ell}}$ (GP RH, surface case). $\textbf{RH}_{\ell}$ and $\textbf{GP}_{\ell}$ use a unitary scale factor. $\textbf{RH}_{\mathring{\ell}}$ uses a random scale factor with mean $1$ and variance $10^{-4}$; $\textbf{GP}_{\mathring{\ell}}$ uses a random scale factor with mean $2/3$ and variance $1/18$.
	
	All estimators are initialized with the true position and velocity vector. FL:OREO estimators are initialized with a uniform class distribution, while the RH estimators are initialized with a circular contour with a radius of $10\,[\mathrm{m}]$.

$\textbf{RH}_{\ell}$ and $\textbf{RH}_{\mathring{\ell}}$ employ $11$ Fourier coefficients; $\textbf{GP}_{\ell}$ and $\textbf{GP}_{\mathring{\ell}}$ discretize the radial function at the equally spaced angles  $\{\frac{2\pi i}{55}\}_{i=1}^{55}$. The covariance matrices of the FL:OREO estimators are respectively
	\begin{align}
	Q^{\textbf{T2S}}&\triangleq \mathrm{diag}\left(\!\!
	\begin{array}{c}
	1, \\
	\text{$[\mathrm{m}]$}
	\end{array}\!\!\!
	\begin{array}{c}
	1, \\
	\text{$[\mathrm{m}]$}
	\end{array}\!\!\!
	\begin{array}{c}
	0.05, \\
	\text{$[\mathrm{rad}]$}
	\end{array}\!\!\!
	\begin{array}{c}
	10, \\
	\text{$[\mathrm{m}/\mathrm{s}]$}
	\end{array}\!\!\!
	\begin{array}{c}
	0.1, \\
	\text{$[\mathrm{m}/\mathrm{s}^2]$}
	\end{array}\!\!\!
	\begin{array}{c}
	0.1 \\
	\text{$[\mathrm{rad}/\mathrm{s}]$}
	\end{array}
	\!\!\right)^2\\
	E^{\textbf{T2S}}&\triangleq \mathrm{diag}\left(10\,[\mathrm{m}],10\,[\mathrm{m}], 5\,[\mathrm{rad}]\right)^2.
	\end{align}
	Moreover, the FL:OREO estimators use the following shaping parameters: $\tau\triangleq 2\,[\mathrm{m}^{-1}]$ (contour case), ${\tau}\triangleq 2\,[\mathrm{m}^{-2}]$ (surface case), $\Delta R \triangleq I\,[\mathrm{m}^2]$, $N\triangleq 100$.
	$\textbf{RH}_{\ell}$ and $\textbf{RH}_{\mathring{\ell}}$ use the process noise covariance
	\begin{equation}
	Q^{\textbf{RH}}\triangleq \mathrm{diag}\left(\!\!
	\begin{array}{c}
	0.1\,I, \\
	\text{$[\mathrm{m}]$}
	\end{array}\!\!\!
	\begin{array}{c}
	1, \\
	\text{$[\mathrm{m}]$}
	\end{array}\!\!\!
	\begin{array}{c}
	1, \\
	\text{$[\mathrm{m}]$}
	\end{array}\!\!\!
	\begin{array}{c}
	10, \\
	\text{$[\mathrm{m}/\mathrm{s}]$}
	\end{array}\!\!\!
	\begin{array}{c}
	10 \\
	\text{$[\mathrm{m}/\mathrm{s}]$}
	\end{array}
	\!\!\right)^2
	\end{equation}
	where the identity matrix refers to estimation of the Fourier coefficients. $\textbf{GP}_{\ell}$ and $\textbf{GP}_{\mathring{\ell}}$ consider the process noise covariance $Q^{\textbf{GP}}\triangleq \mathrm{diag}\left(\overline{Q},Q^{\textbf{f}}\right)$, with
	\begin{equation}
	\overline{Q} \triangleq 
	\left[\begin{array}{cc}
	\frac{T^3}{3} & \frac{T^2}{2} \\[0.05cm]
	\frac{T}{2}   & T
	\end{array}\right]
	\!\otimes \,\mathrm{diag}\left(\begin{array}{c}
	100, \\
	\text{$[\mathrm{m}^2/\mathrm{s}^3]$}
	\end{array}\!\!\!\begin{array}{c}
	100, \\
	\text{$[\mathrm{m}^2/\mathrm{s}^3]$}
	\end{array}\!\!\!
	\begin{array}{c}
	0.02 \\
	\text{$[\mathrm{rad}^2/\mathrm{s}^3]$}
	\end{array}
	\right)
	\end{equation}
	and $Q^{\textbf{f}}$ set as in \cite{wahlstrom2015GP}. All estimators consider the true noise covariance $R$.

	In order to compare the accuracy of the estimators, three different metrics are considered. The first metric is the \textit{Normalized Position Error} (NPE), defined as
	\begin{equation}
	\label{eq:posError}
	\mathrm{NPE}_k \triangleq \frac{\lVert g_k-g_{k|k}\rVert}{\rho_{\min}} \qquad k=1,2,3,\dots
	\end{equation}
	where $g_k$ denotes the true position (i.e. the barycenter) at time $t=(k-1)\,T$ and $g_{k|k}$ is its estimate. The second metric is the \textit{Intersection Over (the) Union} (IOU),
	\begin{equation}
	\label{eq:IOU}
	\mathrm{IOU}_k \triangleq \frac{\left|\mathring{\ell}_k\cap \mathring{\ell}_{k|k}\right|}{\left|\mathring{\ell}_k\cup \mathring{\ell}_{k|k}\right|}
	\qquad k=1,2,3,\dots
	\end{equation}
	where $\ell_k$ denotes the true contour at time $t=(k-1)\,T$ and $\ell_{k|k}$ is its estimate. The \textit{CHamfer Distance} (CHD) \cite{Achlioptas2017LearningRA}, which, for all $k=1,2,\dots$, is defined as
	\begin{align}
	\label{eq:CHD}
	\mathrm{CHD}_k &\triangleq \frac{\mathrm{FPS}\left(\ell_k,\ell_{k|k}\right)+
	\mathrm{FPS}\left(\ell_{k|k}, \ell_k\right)
	}{2} \\
	\label{eq:DCHD}
	\mathrm{FPS}\left(\ell_1,\ell_2\right)&\triangleq \frac{1}{|\ell_1|}\int_{\ell_1} \min_{\zeta \in \ell_2} \lVert z-\zeta\rVert\,\mathrm{d}z.
	\end{align}
 
	is the third and final metric. Compared to the IOU, the CHD penalizes more severely the shape error (intended, for example, as 
	$\int_0^1 \lVert \overline{\psi}_k(\alpha)-\overline{\psi}_{k|k}(\alpha)\rVert\,\mathrm{d}\alpha$), while is more tolerant with respect to the position and orientation errors.
	To accomodate (\ref{eq:IOU}) and (\ref{eq:CHD}), the contours ${c}_{k|k}$ produced by $\textbf{RH}_{\ell}, \textbf{RH}_{\mathring{\ell}}, \textbf{GP}_{\ell}, \textbf{GP}_{\mathring{\ell}}$ are approximated with polygonal lines $\ell_{k|k}$ having $55$ vertices. Such vertices are obtained by sampling the radial functions at the equally spaced angles $\{2\pi\frac{i-1}{54}\}_{i=1}^{55}$. To accomodate (\ref{eq:posError}), the position estimates generated by $\textbf{RH}_{\ell}, \textbf{RH}_{\mathring{\ell}}, \textbf{GP}_{\ell}, \textbf{GP}_{\mathring{\ell}}$ are defined as the barycenters $g_{k|k}$ of the relative contour estimates $\ell_{k|k}$. 
	
In terms of the proposed metrics, FL:OREO estimators outperform the RH and GP estimators. This because FL:OREO estimators, thanks to the dictionary, have a significant advantage regarding shape estimation in terms of prior information.
Indeed, since the dictionary contains the true shape of the tracked object, once the true class is identified, the shape error generated by FL:OREO estimators become exactly zero. On the other hand, RH and GP are designed to work without such prior information, and so that cannot achieve a null shape estimation error.

%In particular, all versions of FL:OREO are essentially equivalent due to the fact that such estimators, despite the convergence rates of the respective classifiers are different, only in few initial scans (typically, less than 5) estimates wrongly the shape of the object.

The computational times per single scan of FL:OREO estimators are considerably larger than the ones of RH and GP estimators. This is a consequence of the fact that FL:OREO, unlike RH and GP, evaluates $\mathcal{I}$ different correction models at each sensor scan.

\begin{figure}
	\centering
	\includegraphics[scale=0.35]{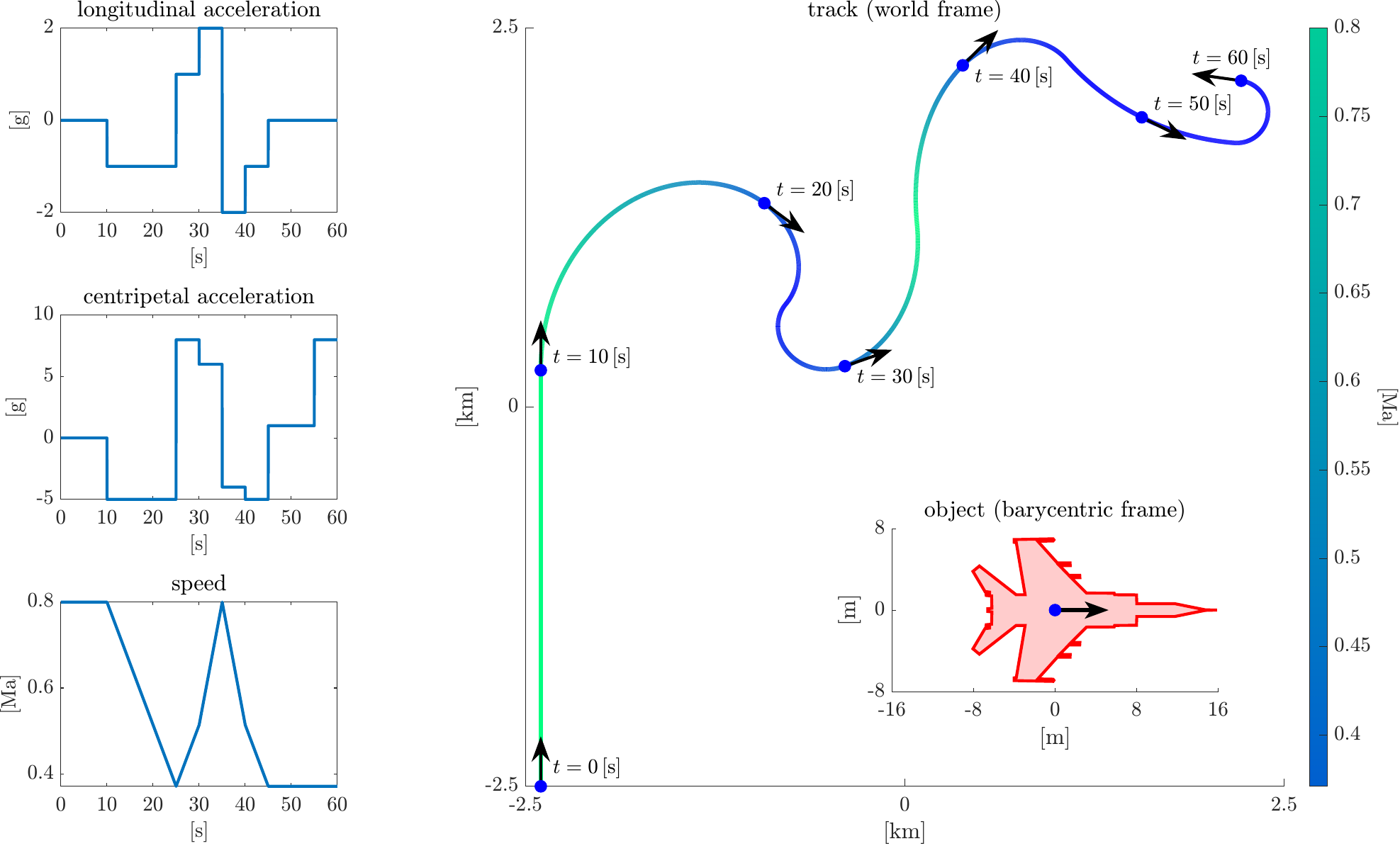}
	\caption{ground truth of the tracking test. Driving inputs are expressed in $\mathrm{g}\approx 9.81\,[\mathrm{m}/\mathrm{s}^{2}]$ and $\mathrm{Ma}\approx 343\,[\mathrm{m}/\mathrm{s}]$ units.
	}
	\end{figure}
		
	\begin{figure}
	\centering
	\includegraphics[scale=0.25]{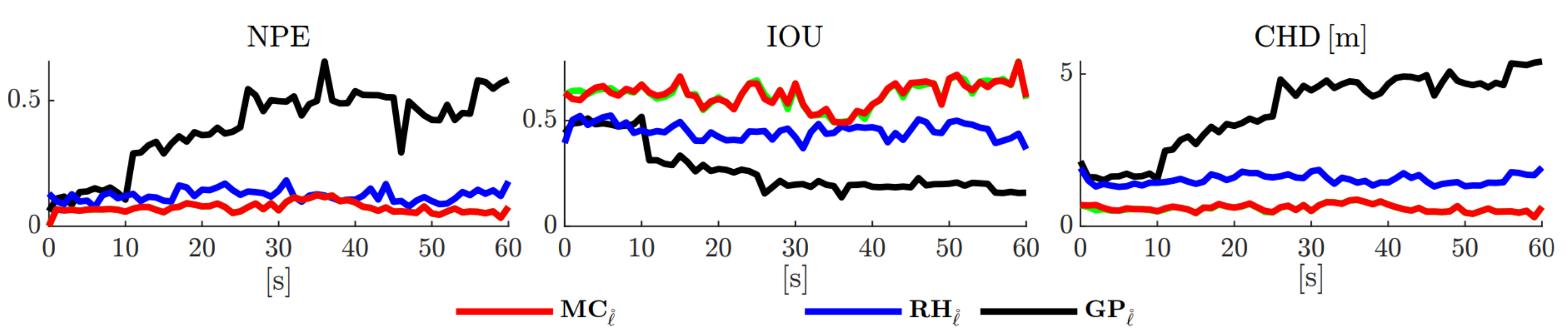}
	\caption{estimation scores of the specific tracking test $\mathring{\overline{r}}\triangleq 10\,\,[\mathrm{m}^2]$, $\sigma\triangleq 1\,\,[\mathrm{m}]$. The results are averaged over $10^2$ runs.
	}
	\end{figure}

	\begin{table*}
    \centering
    \ra{0.8}
    
    \begin{tabular}{rccccccccc}
        \toprule
          & \multicolumn{3}{c}{$\overline{r} \triangleq 1 \,\left[\mathrm{m}\right]$\hspace{0.5cm}($\overline{m}_{\ell}\approx86$)} &
          & \multicolumn{3}{c}{$\overline{r} \triangleq 10 \,\left[\mathrm{m}\right]$\hspace{0.5cm}($\overline{m}_{\ell}\approx9$)}\\
        \cmidrule{2-4} \cmidrule{6-8} 
          & \multicolumn{3}{c}{$\sigma\triangleq 0.1\,[\mathrm{m}]$} &
          & \multicolumn{3}{c}{$\sigma\triangleq 1\,[\mathrm{m}]$} \\
        \cmidrule{2-4} \cmidrule{6-8}
          & $\mathrm{NPE}$ 
          & $\mathrm{IOU}$ 
          & $\mathrm{CHD}\,\,[\mathrm{m}]$ & 
          & $\mathrm{NPE}$ 
          & $\mathrm{IOU}$ 
          & $\mathrm{CHD}\,\,[\mathrm{m}]$ \\
        \midrule
        
        $\textbf{EC}\!\phantom{00}$     
                            & $0.07$ & $0.66$ & $0.51$ & 
                            & $0.07$ & $0.66$ & $0.51$ \\
        $\textbf{RH}_{\ell}\phantom{0}$     
                            & $0.21$ & $0.29$ & $1.68$ & 
                            & $0.07$ & $0.60$ & $0.93$ \\
        $\textbf{GP}_{\ell}\phantom{0}$     
                            & $0.23$ & $0.41$ & $0.66$ & 
                            & $0.16$ & $0.48$ & $1.59$ \\
        \midrule
          & \multicolumn{3}{c}{$\mathring{\overline{r}} \triangleq 1 \,\left[\mathrm{m}^2\right]$\hspace{0.5cm}($\overline{m}_{\mathring{\ell}}\approx95$)} &
          & \multicolumn{3}{c}{$\mathring{\overline{r}} \triangleq 10 \,\left[\mathrm{m}^2\right]$\hspace{0.5cm}($\overline{m}_{\mathring{\ell}}\approx10$)}\\
        \midrule
        $\textbf{MC}\!\phantom{00}$     
                            & $0.04$ & $0.76$ & $0.34$ & 
                            & $0.05$ & $0.75$ & $0.35$ \\
        $\textbf{RH}_{\mathring{\ell}}\phantom{0}$     
                            & $0.21$ & $0.25$ & $1.94$ & 
                            & $0.05$ & $0.53$ & $1.27$ \\
        $\textbf{GP}_{\mathring{\ell}}\phantom{0}$     
                            & $1.03$ & $0.16$ & $8.09$ & 
                            & $0.46$ & $0.24$ & $4.93$ \\
        \bottomrule
    \end{tabular}
    
\end{table*}

\begin{table*}
    \centering
    \ra{0.8}
    
    \begin{tabular}{rccccccccc}
        \toprule
          & \multicolumn{3}{c}{$\overline{r} \triangleq 1 \,\left[\mathrm{m}\right]$\hspace{0.5cm}($\overline{m}_{\ell}\approx86$)} &
          & \multicolumn{3}{c}{$\overline{r} \triangleq 10 \,\left[\mathrm{m}\right]$\hspace{0.5cm}($\overline{m}_{\ell}\approx9$)}\\
        \cmidrule{2-4} \cmidrule{6-8} 
          & \multicolumn{3}{c}{$\sigma\triangleq 0.1\,[\mathrm{m}]$} &
          & \multicolumn{3}{c}{$\sigma\triangleq 1\,[\mathrm{m}]$} \\
        \cmidrule{2-4} \cmidrule{6-8}
          & $\mathrm{NPE}$ 
          & $\mathrm{IOU}$ 
          & $\mathrm{CHD}\,\,[\mathrm{m}]$ & 
          & $\mathrm{NPE}$ 
          & $\mathrm{IOU}$ 
          & $\mathrm{CHD}\,\,[\mathrm{m}]$ \\
        \midrule
        
        $\textbf{EC}\!\phantom{00}$     
                            & $0.11$ & $0.55$ & $0.77$ & 
                            & $0.10$ & $0.55$ & $0.76$ \\
        $\textbf{RH}_{\ell}\phantom{0}$     
                            & $0.27$ & $0.23$ & $1.96$ & 
                            & $0.13$ & $0.50$ & $1.37$ \\
        $\textbf{GP}_{\ell}\phantom{0}$     
                            & $0.24$ & $0.38$ & $2.29$ & 
                            & $0.23$ & $0.39$ & $2.11$ \\
        \midrule
          & \multicolumn{3}{c}{$\mathring{\overline{r}} \triangleq 1 \,\left[\mathrm{m}^2\right]$\hspace{0.5cm}($\overline{m}_{\mathring{\ell}}\approx95$)} &
          & \multicolumn{3}{c}{$\mathring{\overline{r}} \triangleq 10 \,\left[\mathrm{m}^2\right]$\hspace{0.5cm}($\overline{m}_{\mathring{\ell}}\approx10$)}\\
        \midrule
        $\textbf{MC}\!\phantom{00}$     
                            & $0.07$ & $0.63$ & $0.58$ & 
                            & $0.07$ & $0.62$ & $0.60$ \\
        $\textbf{RH}_{\mathring{\ell}}\phantom{0}$     
                            & $0.22$ & $0.24$ & $1.86$ & 
                            & $0.12$ & $0.45$ & $1.54$ \\
        $\textbf{GP}_{\mathring{\ell}}\phantom{0}$     
                            & $0.95$ & $0.17$ & $9.25$ & 
                            & $0.39$ & $0.26$ & $3.80$ \\
        \bottomrule
    \end{tabular}
    
    \caption{time-averaged estimation scores. The results are averaged over $10^2$  runs.}
\end{table*}

	\begin{table*}
	
	\centering
	\ra{0.8}
	
	\begin{tabular}{rrrrr}
	\toprule
	
	 & $\textbf{MC}\phantom{00}$
	 & $\textbf{EC}\phantom{00}$        
	 & $\textbf{RH}\phantom{00}$ 
	 & $\textbf{GP}\phantom{00}$ \\
	 \midrule
	$m\triangleq 10^0$ 
							& $36\,[\mathrm{ms}]$ 
	                & $24\,[\mathrm{ms}]$  & $2\,[\mathrm{ms}]$ 
	                & $2\,[\mathrm{ms}]$ \\
	$m\triangleq 10^1$ 
							& $40\,[\mathrm{ms}]$ 
	                & $51\,[\mathrm{ms}]$  & $8\,[\mathrm{ms}]$ 
	                & $16\,[\mathrm{ms}]$ \\
	$m\triangleq 10^2$ 
							& $127\,[\mathrm{ms}]$ 	                 
	                & $275\,[\mathrm{ms}]$  & $60\,[\mathrm{ms}]$ 
	                & $105\,[\mathrm{ms}]$ \\
	$m\triangleq 10^3$ 
							& $1144\,[\mathrm{ms}]$ 	                 
	                & $2531\,[\mathrm{ms}]$  & $525\,[\mathrm{ms}]$ 
	                & $4840\,[\mathrm{ms}]$ \\                
	\bottomrule
	\end{tabular}
	
	\caption{time-averaged execution times per single scan. The results are averaged over $10^2$ runs.
	The results have been generated by a common laptop (Intel
	 i5-1135G7) exploiting a MATLAB \texttt{parpool} composed by 4 workers.}
	\end{table*}

 \section{Conclusions and future developments}
	Two likelihood models for extended objects have been developed (Monte Carlo, Exact Contour). Such models allows to effectively solve tracking and classification problems involving extended objects whose shape can be arbitrarily complex. Numerical simulations provide an empirical evidence that the proposed estimators are more accurate and computationally intensive than state of the art estimators.  
 %Moreover, (\ref{eq:generalLik}), the proposed samplers and the proposed generators show how the linear spline framework opens up several research directions concerning extended object tracking and classification. 
	
	At the single object level, the following challenging problems will be investigated: 3-dimensional tracking based on polyhedra; model-based tracking of stealth objects; tracking of self-occluded objects.
	In addition to these topics, an \textit{agnostic} extension, i.e. a dictionary-free solution, will be investigated. 
	
	More interestingly, at the multi-object level, the following challenging problems will be investigated:
	multi-object, multi-sensor and multi-scan tracking of extended objects in presence of clutter, self-occlusions and proper occlusions. To achieve this goal, the proposed methods will be cast in state of the art multi-object filters based on RFSs, such as \textit{Generalized Labeled Multi-Bernoulli} filters \cite{vo2013GLMB1, vo2014GLMB2, vo2016GLMB3}.

\bibliographystyle{unsrt}  
\bibliography{references}  %%% Remove comment to use the external .bib file (using bibtex).

\end{document}